\documentclass[conference]{IEEEtran}
\IEEEoverridecommandlockouts
\usepackage{cite}
\usepackage{amsmath,amssymb,amsfonts}
\usepackage{algorithmic}
\usepackage{graphicx}
\usepackage{textcomp}
\usepackage{subcaption}
\usepackage{xcolor}
\usepackage[ruled,vlined]{algorithm2e}
\def\BibTeX{{\rm B\kern-.05em{\sc i\kern-.025em b}\kern-.08em
    T\kern-.1667em\lower.7ex\hbox{E}\kern-.125emX}}
\begin{document}

\title{Prognostics-Informed Battery Reconfiguration in a Multi-Battery Small UAS Energy System\\
\thanks{This research was supported in part by National Science Foundation I/UCRC grant 1738714, the Center for Unmanned Aircraft Systems (C-UAS).}
}

\author{\IEEEauthorblockN{Prashin Sharma}
\IEEEauthorblockA{\textit{Robotics Institute} \\
\textit{University of Michigan}\\
Ann Arbor, USA \\
prashinr@umich.edu}
\and
\IEEEauthorblockN{Ella Atkins}
\IEEEauthorblockA{\textit{Aerospace Engineering and Robotics Institute} \\
\textit{University of Michigan}\\
Ann Arbor, USA \\
ematkins@umich.edu }
}

\maketitle

\begin{abstract}
Batteries have been identified as one most likely small UAS (sUAS) components to fail in flight. sUAS safety will  therefore be improved with redundant or backup batteries. This paper presents a prognostics-informed Markov Decision Process (MDP) model for managing multi-battery reconfiguration for sUAS missions. Typical lithium polymer (Lipo) battery properties are experimentally characterized and used in Monte Carlo simulations to establish battery dynamics in sUAS flights of varying duration. Case studies illustrate the trade off between multi-battery system increased complexity/weight and resilience to non-ideal battery performance.  
\end{abstract}


\section{Introduction}
Multicopters are a popular platform for emerging low-altitude small unmanned aircraft system (sUAS) missions such as inspection, surveillance, and package delivery \cite{Wing, Flytrex, Zipline}. However, risk to an overflown population can be nontrivial due to uncertainty in actuator and battery performance, external disturbances, lost link, and more.  In a survey conducted of 1500 companies using drones in the UK \cite{Osborne2019} the average sUAS failure rate is $10^{-3}$ per flight hour. Multicopter battery and motor failures are particularly problematic because energy and thrust margins are limited. 

Prognostics methods provide tools for predicting failure scenarios and sending key health updates enabling updates to ensure an sUAS can safely reach its destination. This paper presents an experimentally-derived battery model for standard lithium-polymer (Lipo) multicopter batteries and uses this model in a Markov Decision Process (MDP) based battery management framework.  This work assumes a multicopter sUAS carries multiple battery packs because a single pack does not offer backup should a failure (e.g., internal short, failed cell) occur.  

This paper offers the following contributions:
\begin{itemize}
    \item An experimentally-validated sUAS battery model is presented and embedded in sUAS Monte Carlo simulations.
    \item An MDP for sUAS battery management is proposed.  
    \item Case studies examine battery and MDP performance over a variety of different health and mission length scenarios.
\end{itemize}

The remainder of the paper is organized as follows. A literature review is followed by a problem statement in Section \ref{Sec:prob}. Section \ref{Sec:model} presents a parameter identification process and an experimentally-derived sUAS battery model.  A motor current draw model and battery management MDP are also defined.  Section \ref{Sec:MCSim} describes Monte Carlo simulations used to determine MDP state transition probabilities.  Results and conclusions are presented in Sections \ref{Sec:Results} and \ref{Sec:Conclusion}, respectively.

\section{Literature Review}

Prognosis algorithms are critical for sUAS safety assurance. In \cite{chetan2019health}, the author presents a methodology for health management in an electric unmanned air vehicle (UAV). The probability of subsystems failure is quantified by combining failure mode, effects, and critical analysis using a qualitative Bayesian approach. A distributed diagnosis algorithm detects and diagnoses a failed power train subsystem and instantiates a prognoser to determine remaining useful life (RUL) of the faulty system. A model-based prognosis algorithm is proposed \cite{schacht2018prognosis} to determine remaining flight time for an electric UAV. The authors use an Extended Kalman Filter (EKF) to estimate lithium polymer battery state of charge (SoC). A polynomial function is used to estimate SoC until the low voltage threshold is reached to determine flight endurance, assuming no sudden variation in SoC. In \cite{hogge2018verification}, the authors  use an Unscented Kalman Filter (UKF) to determine battery SoC. Assuming the remaining flight plan was known, the UKF was used to simulate the flight plan energy demand into the future until an critical SoC threshold was reached. These approaches have applied model-based methods for prognostics. In \cite{ELEFTHEROGLOU2019} the authors use a data-driven strategy to determine RUL and predict battery end of discharge (EOD) time dynamically. Machine learning algorithms such as Gradient Boost Trees (GBT), Bayesian Neural Network and Non-Homogeneous Hidden Semi Markov Model (NHHSMM) have been proposed. Ref. \cite{ELEFTHEROGLOU2019} recommends NHHSMM and GBT over Bayesian Neural Networks due to potential model constraint violations and difficulty closing the distance between upper and lower RUL estimate bounds with the neural network approach.

Prognostics information alone is not very helpful unless used in an active system for preventative maintenance or contingency management.  Ref. \cite{balaban2013modeling} proposes the Prognostics-based Decision Making (PDM) architecture consisting of four main elements: a diagnoser (DX), decision maker (DM), vehicle simulation(VS) and the vehicle itself. The prognostics problem is formulated as a constraint satisfaction problem (CSP) and solved using backtracking search and particle filtering. In this framework, mission waypoints are defined a-priori; waypoints are assumed reachable even in the presence of faults. A similar prognostics architecture is proposed in \cite{balaban2013development} and implemented on an unmanned ground vehicle. In \cite{schacht2019} the authors proposed a mission planning strategy for sUAS multicopters that incorporates battery State of Charge (SoC) and State of Health (SoH) to generate updated mission plans. The planning problem is formulated as an optimization problem to minimize total energy consumed by the multicopter, subject to nonlinear constraints defined by UAV dynamics, brushless motor dynamics and battery dynamics.  In reference \cite{tang2008} and follow-on work \cite{tang2010} the authors present an Automated Contingency Planner enhanced by prognostic information. Online optimization determines a minimum cost reconfiguration for the system and components. A receding horizon planner is utilized in \cite{zhang2014} to incorporate the constraints determined from prognostics information. 

Ref. \cite{Poteiger2017} utilizes prognostics outputs in battery reconfiguration using a constraint satisfaction algorithm. A survey of battery reconfiguration techniques is presented in \cite{Ci2016}. In \cite{Jain2020}, the authors present a novel method for in-flight battery swap. A primary battery is consistently carried onboard a long-endurance host drone, and a smaller drone carrying a secondary battery sporadically attaches to the primary or host drone. Once docked, the host drone switches from primary to secondary battery, and once the secondary battery depletes, the smaller "recharging" drone undocks from the host drone. 

In \cite{IVERSEN20141} the authors proposes an algorithm to optimally charge electric vehicles. Hidden Markov Models (HMMs) are obtained from fitting data collected from stochastic usage patterns for a single vehicle. Dynamic programming is used to determine optimal charge policies. There is no mention of battery health in this work.
Life-cycle assessment for Lipo batteries using a Markov Decision Process (MDP) is proposed in \cite{THEIN2014142}. Optimal policies obtained from solving the MDP act as a condition-based decision-maker to either recycle, inspect, or reuse a battery.

Our proposed method's innovations relative to state-of-the-art are in the specification and evaluation of a prognostics-informed MDP for battery management in multi-battery sUAS.

\section{Problem Statement}
\label{Sec:prob}

\begin{figure}
        \centering
        \includegraphics[width = 2in , height=2.25in]{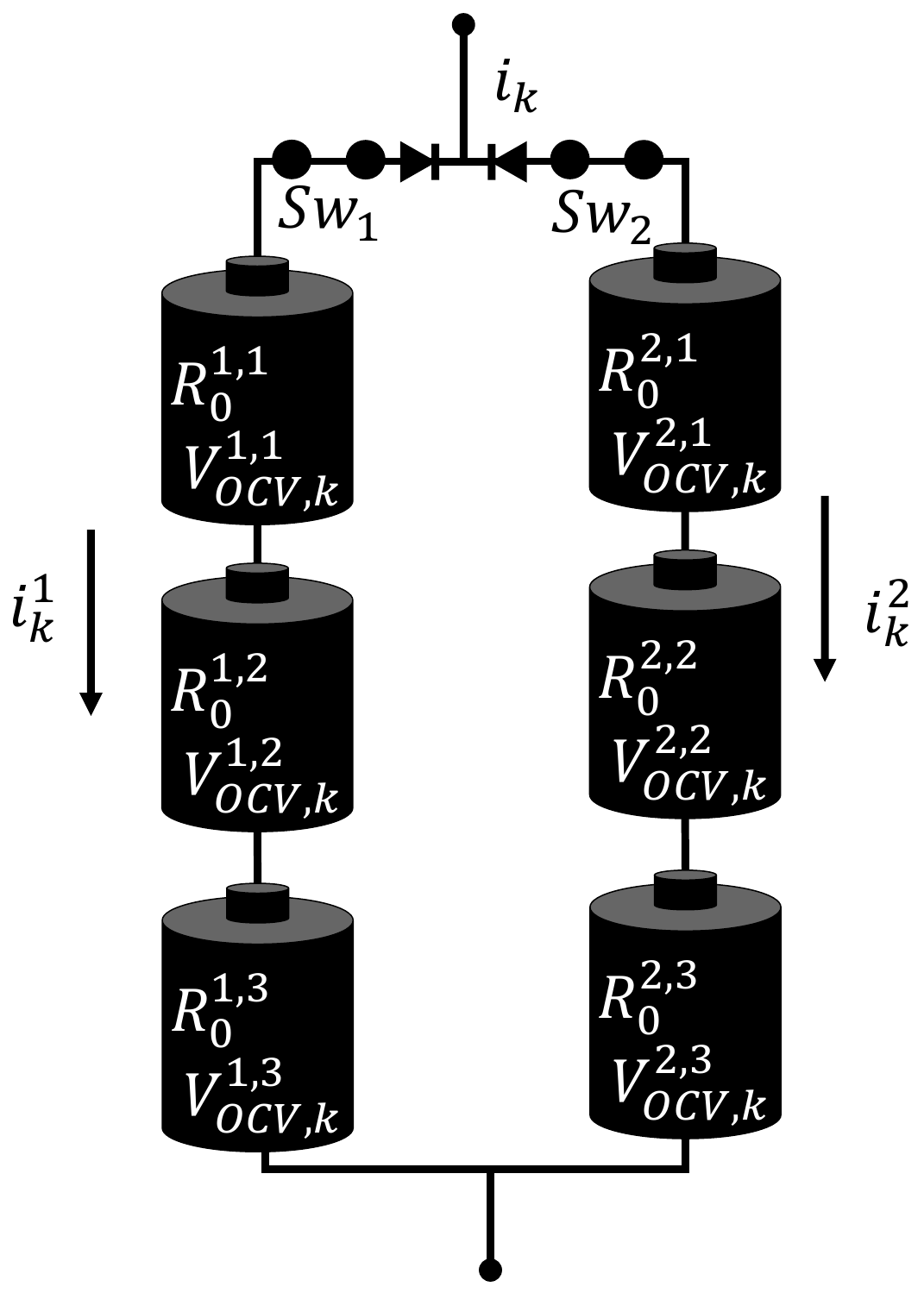}
        \caption{The sUAS series-parallel battery pack system modeled in this work.}
        \label{fig:series-parallel}
    \end{figure}
\begin{figure}[!htbp]
  \centering
    \includegraphics[width=0.35\textwidth]{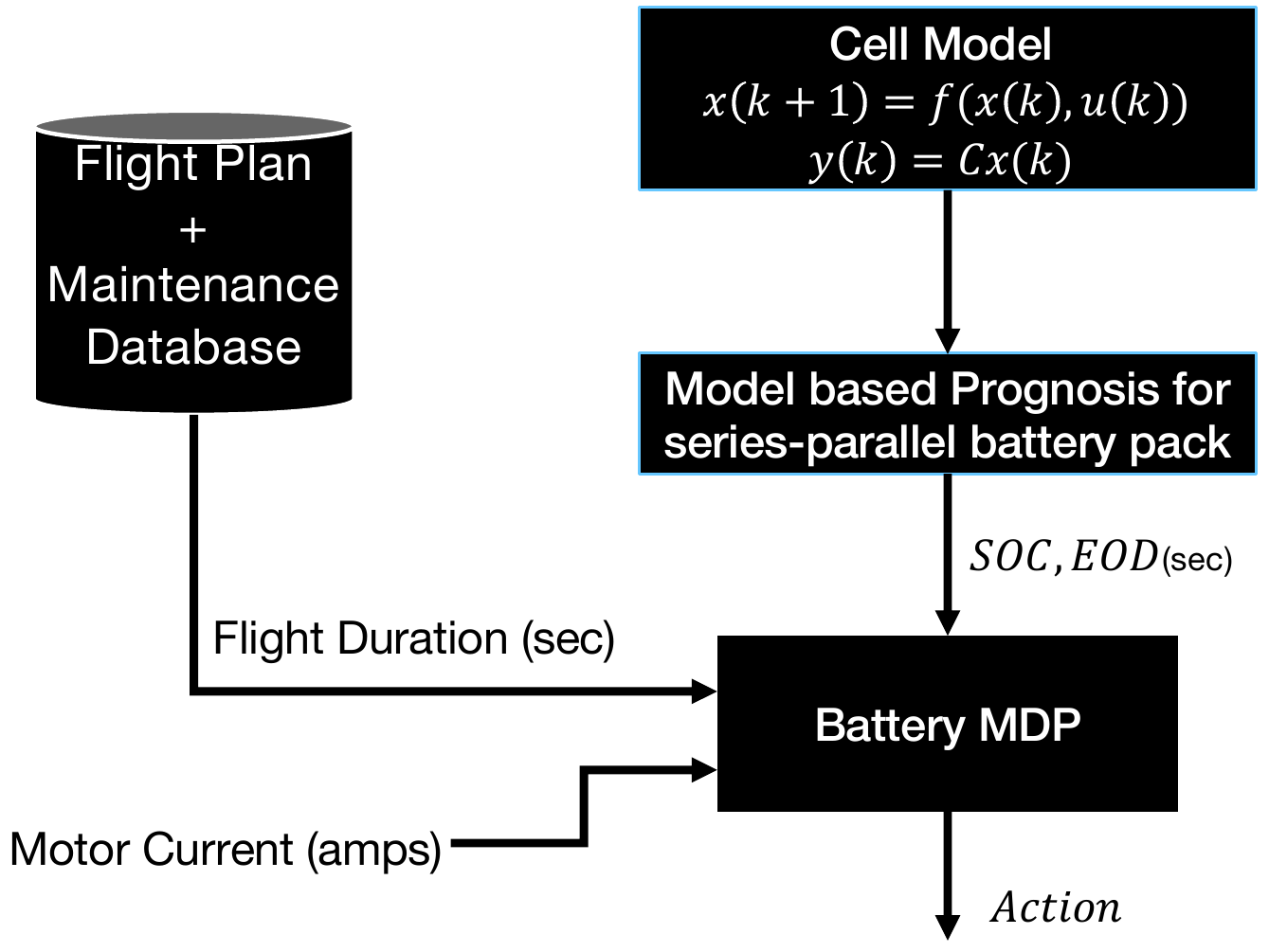}
    \caption{Battery reconfiguration Markov Decision Process (MDP) Model.}
    \label{fig:BattMDP}
\end{figure}

Battery failures are one of the top three causes of sUAS failure.  Series-parallel battery pack capacity fade and elevated impedance typically cause imbalance resulting in permanent battery pack degradation \cite{Mikolajczak2011}. Ref. \cite{Michael2018} recommends battery failures be mitigated by employing batteries in parallel.  
This paper proposes a battery reconfiguration Markov Decision Process (MDP) for a series-parallel battery pack as shown in Figure \ref{fig:series-parallel}. This MDP considers battery pack degradation conditions and remaining flight time in recommending battery use and switching decisions.  The proposed battery MDP architecture is shown in Figure \ref{fig:BattMDP}. The MDP reconfiguration concept was first proposed in the context of a mission planning hierarchical MDP (hMDP) in \cite{Prashin2020}. This paper presents the battery and MDP models, simulations, and case study results associated with the battery MDP component of the previously-proposed hMDP architecture.
As shown in the figure, the battery MDP outputs an optimal battery usage action based on remaining flight time, expected motor current draw, and each battery's End of Discharge (EoD) time. Accurate models of battery cell and pack charge and discharge dynamics, mission execution including expected motor current draw and remaining flight time, and battery health must be developed.  The MDP action and state-space must be decomposed to the extent possible to minimize computational overhead and facilitate explainability.  Rewards must reflect user and mission preferences, and state transition probabilities must be obtained from realistic Monte Carlo simulations.

\section{System Modelling}
\label{Sec:model}
\subsection{Enhanced Self-Correcting Battery Model}
\begin{figure}[!htbp]
  \centering
    \includegraphics[width=0.3\textwidth]{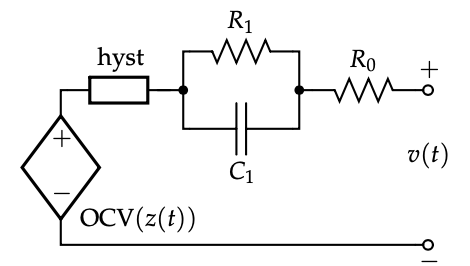}
    \caption{Enhanced Self-Correcting System (ECS) model of a Lipo Battery.}
    \label{fig:ECR}
\end{figure}

The Lipo battery can be modelled as a voltage source (Open Circuit Voltage) that has a resistance and one or more sub-circuits of resistance and capacitor in parallel with hysteresis as shown in Figure \ref{fig:ECR}. This model is referred to as the Enhanced Self-Correcting System (ECS) Model.

The discrete state space ECS model formulation \cite{plett2015battery} using multiple parallel resistor-capacitor pair current can be written as:
\begin{equation}\label{eq:IR}
    i_r[k+1] = 
    \underbrace{\begin{bmatrix}
    F_1 & 0 & \cdots \\
    0 & F_2 & \\
    \vdots & & \ddots
    \end{bmatrix}}_{A_{RC}} i_r[k] + \underbrace{\begin{bmatrix}
    (1- F_1) \\ (1-F_2) \\ \vdots 
    \end{bmatrix}}_{B_{RC}}i[k]
\end{equation}
where $i_r$ is the current through the sub-circuit of the parallel resistor-capacitor and $ F_j = exp(\frac{-\Delta t}{R_j C_j})$. 
Let $A_H[k] = exp(-|\frac{\eta[k] i[k] \gamma \Delta t}{Q}|) $. The state-space equation for the ECS model is given by: 

\begin{equation} \label{eq:ESC}
\begin{split}
    \begin{bmatrix} z[k+1] \\ i_r[k+1] \\ h[k+1] \end{bmatrix} =& \begin{bmatrix} 1 & 0 & 0 \\ 0 & A_{RC}  & 0 \\ 0 & 0 & A_H[k] \end{bmatrix}  \begin{bmatrix} z[k] \\ i_r[k] \\ h[k]\end{bmatrix}\\& +  
    \begin{bmatrix} 
        \frac{-\eta[k]\Delta t}{Q} & 0 \\
        B_{RC} & 0 \\
        0 & (A_H[k]-1)
    \end{bmatrix}
    \begin{bmatrix}
        i[k] \\
        sgn(i[k])
    \end{bmatrix}
\end{split}
\end{equation}
where $\eta =$ charge efficiency, $ i[k] =$ current through the circuit, $ \gamma =$ a positive constant, $ Q =$ total charge capacity (ampere-secs),$ z[k] =$ State of Charge (SOC), and $ h[k] =$ dynamic hysteresis. 
The battery voltage output equation is given by:
\begin{equation}
v[k] = OCV(z[k]) + M_0s[k] + Mh[k] + \sum_jR_{j}i_{R_j}[k] - R_0i[k]  
\label{eq:ESCoutput}
\end{equation}
where, $OCV(z[k])=$ Open Circuit Voltage as function of time, $M,M_0 =$ Constants, $s[k] = $Instantaneous hysteresis.

\subsubsection{Battery Model Parameter Identification }
To use the ECS model defined by equations (\ref{eq:IR}), (\ref{eq:ESC}) and (\ref{eq:ESCoutput}) parameters including the OCV vs SOC relation, $\gamma$, $\eta$, $Q$,$R_j$,$R_0$ must be identified. The experimental setup to determine these parameters is described in this section. For all experiments, a 3s standard sUAS Lipo battery was used. 

\begin{figure}
        \centering
        \includegraphics[width=0.35\textwidth]{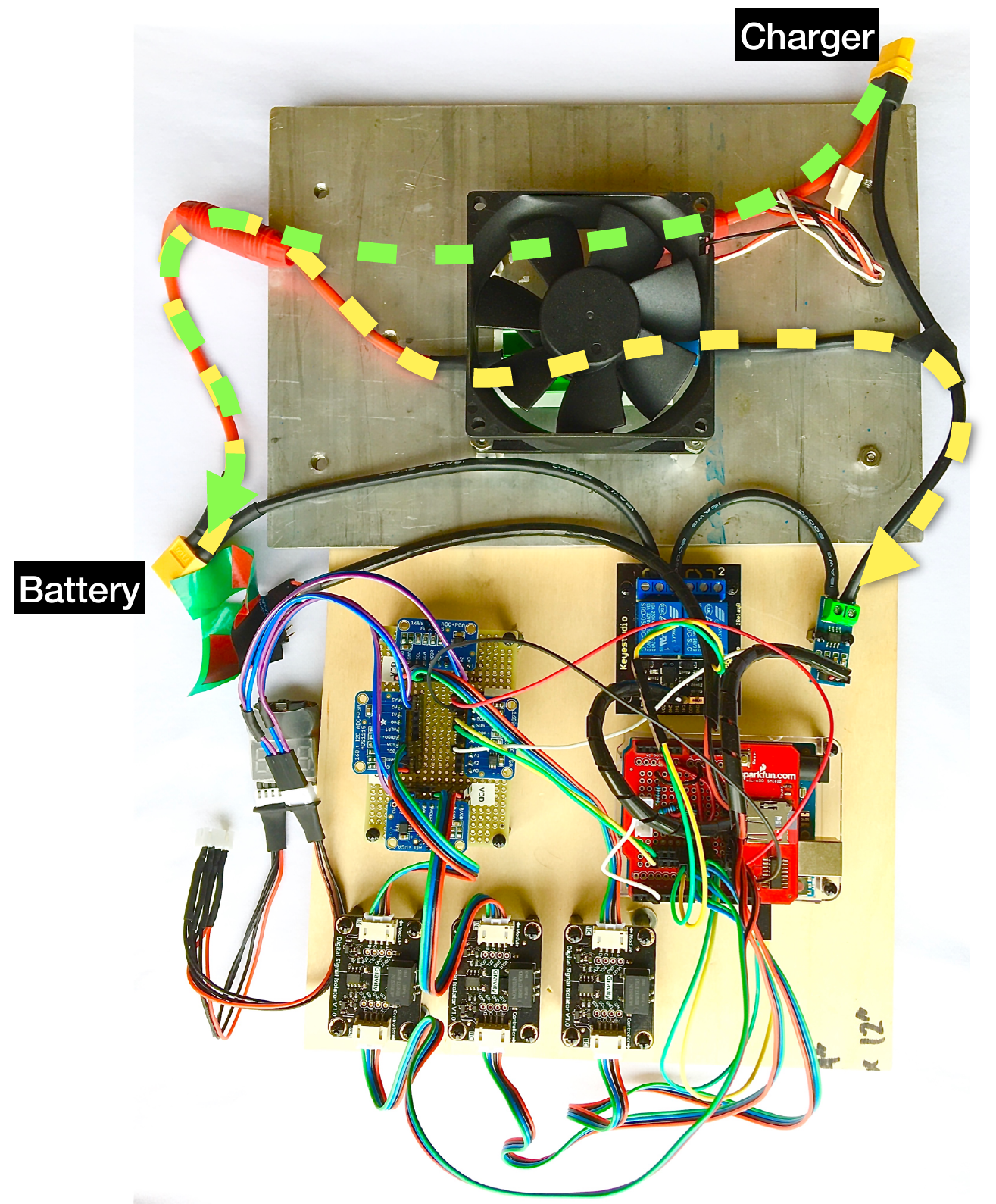}
        \caption{Experimental setup to measure voltage and current during charge/discharges cycle of a 3S Lipo battery.}
        \label{fig:setup}
    \end{figure}

The Lipo battery parameter identification experimental setup  is shown in Figure \ref{fig:setup}. An Arduino UNO was used for data collection. Each of the cell voltages was measured by 16-bit ADC converters \footnote{https://www.adafruit.com/product/1085} connected to analog optoisolators \footnote{https://www.digikey.com/en/products/detail/dfrobot/DFR0504/7682221}. A load of 4$\Omega$ was attached across the battery terminal to draw a maximum current of 3$Amp$ (1C).  Output current was sensed with a low-cost current sensor \footnote{https://www.sparkfun.com/datasheets/BreakoutBoards/0712.pdf} during the discharged cycle. A relay was also added in series with the resistance to turn the resistive load on and off. 
For the battery charge cycle, a standard Lipo charger, a Venom Pro Duo 80W \footnote{https://www.venompower.com/venom-pro-duo-80w-x2-dual-ac-dc-7a-lipo-lihv-nimh-rc-battery-balance-charger-0685} battery balance-charger product, was used. 

\begin{figure}
        \centering
        \includegraphics[width=0.35\textwidth]{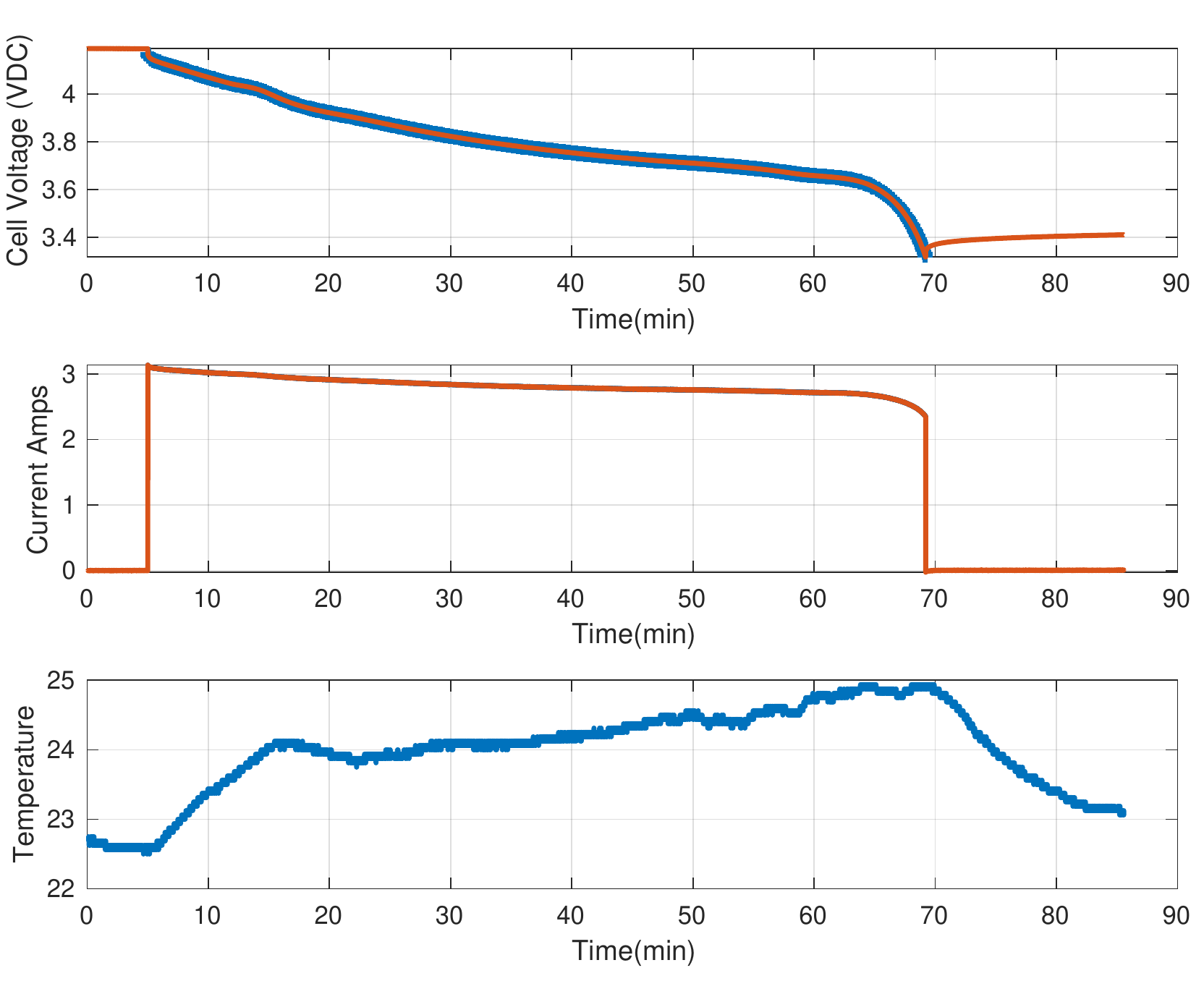}
        \caption{ Time series plot of single cell battery voltage, current and surface temperature($^\circ C$) sensed during a 1C discharge of a 3s battery pack. The blue highlighted segment of cell voltage and current is used for further analysis.}
        \label{fig:discharge}
    \end{figure}

For this experiment, voltage and current measurements over the full charge and discharge cycles of the Lipo battery were collected over time. The battery was discharged to 3.3VDC per cell per the manufacturer's recommendation and charged using the Venom balance-charger. A sample time history of collected data is shown in Figure \ref{fig:discharge}. 
 Because the low-cost battery cells have some manufacturing variability, not all cells reached the set critical voltage limit at the same time; the cells typically differ by a few $mV$. Data was collected at a single room temperature with battery surface temperature variation $\pm 2^\circ C$. The setup can be placed in a temperature-controlled environment to obtain a relationship between parameters and temperature, but this was beyond the scope of our experiments. Our simple setup does not replace manufacturer battery testing systems but does provide an economical option for battery parameter identification and model validation.  

\begin{figure}
    \centering
    \includegraphics[width = 0.35\textwidth]{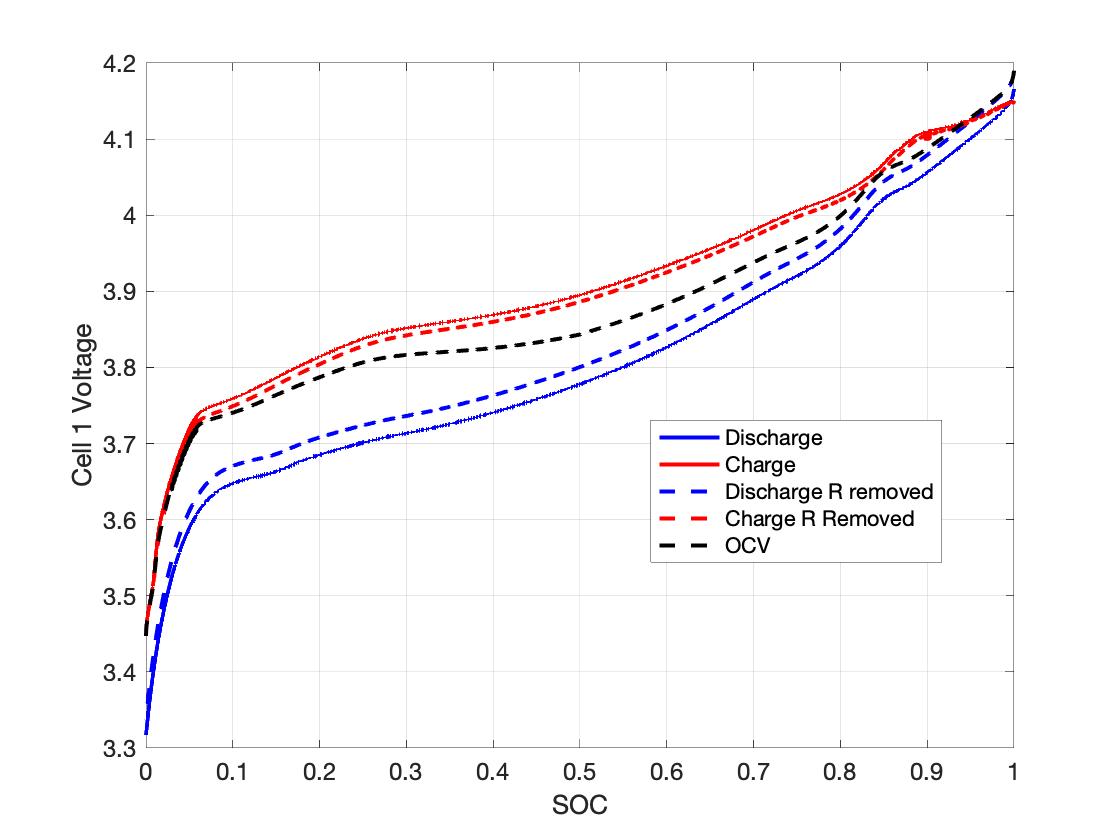}
    \caption{Battery voltage variation with State of Charge (SOC) during the charge/discharge cycle for a single cell.}
    \label{fig:SOC}
\end{figure}

The detailed procedure we applied for parameter identification can be found in \cite{plett2015battery}. The relation between SOC and charge/discharge voltage for each of the cells can be identified, as shown in Figure \ref{fig:SOC}.
After obtaining the charge/discharge relation to SOC we can obtain the OCV relation to SOC. Similar to the approach taken in \cite{saxena2018}, OCV was obtained by taking an average of the charge and discharge curve.

    

\begin{figure}[!htbp]
  \centering
    \includegraphics[width=0.3\textwidth]{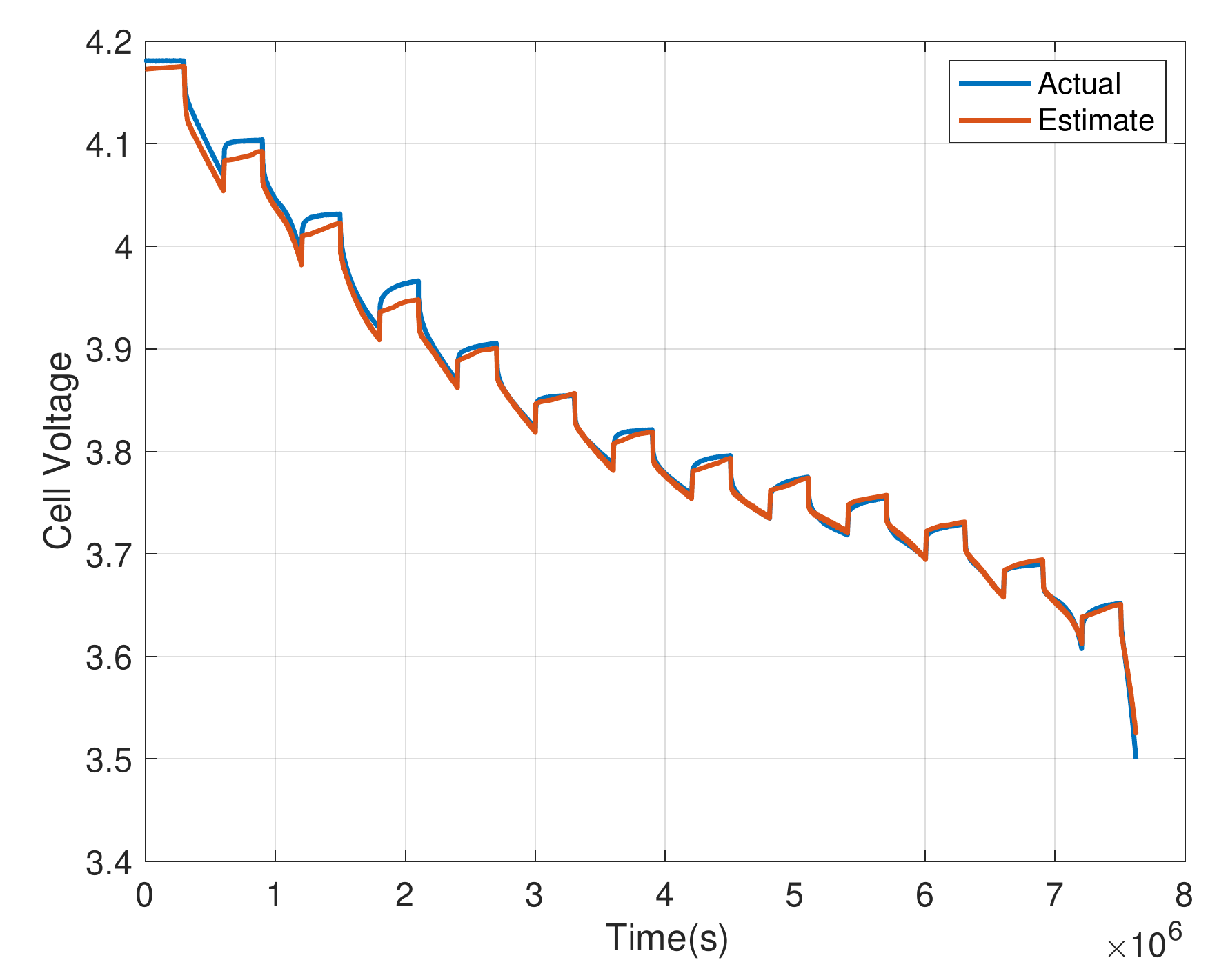}
    \caption{Time series plot comparing model estimate and actual cell voltage measured by the battery data collector during pulse discharge. }
    \label{fig:BattModVal}
\end{figure}

The identified parameters are tabulated in Table \ref{tb:paramH}. 
    \begin{table}[h]
    \caption{Identified parameters for a 3s Lipo battery.}
    \label{tb:paramH}
    \begin{center}
    \begin{tabular}{|c||c||c||c|}
    \hline 
    \textbf{Parameter} & \boldmath{$Cell_1$ }  & \boldmath{$Cell_2$} & \boldmath{$Cell_3$} \\
    \hline
    \textit{$R_0$} ($m\Omega$) &  6 & 5.5 & 5.5  \\
    \hline
    \textit{$R_1$} ($m\Omega$) & 9.9  & 8.8 & 8.7 \\
    \hline
    \textit{$C_1$} ($kF$) & 51.5  & 52.5 & 60.5 \\
    \hline
    \textit{$Q$} ($Ah$) & 3.0271  & 3.0271 & 3.0271 \\
    \hline 
    $\gamma$   &300 & 200 &300 \\ 
    \hline
    \textit{M}($10^{-3}$) & 17.8 & 31.2  & 18.4 \\ 
    \hline
    \textit{$M_0$}($10^{-3}$) & 5.1  & 5.3 & 5.2 \\
    \hline
     $\eta_{Charge}$   & 0.973 & 0.973  & 0.973 \\ 
    \hline
     $\eta_{Discharge}$   & 1 & 1  & 1 \\ 
    \hline
    \end{tabular}
    \end{center}
    \end{table}
 Additional data was collected to validate the identified model. Battery cell response to a pulsed discharge is shown in Figure \ref{fig:BattModVal}. Error was observed on the order of $mV$, which we deemed acceptable for our simulations.  

For simulations the current draw through each branch of the parallel battery pack is shown in Figure \ref{fig:series-parallel} and is determined by the following equations:
\begin{gather}
    V_k = \frac{\sum^{N_P}_{j=1} \frac{V^j_{OCV,k}}{R^j_0} -i_k}{\sum^{N_P}_{j=1} \frac{1}{R^j_0}} \\
    i^j_k =\frac{V^j_{OCV,k} - V_k}{R^j_0}\\
    V^j_{OCV,k} =V^{j,1}_{OCV,k} +V^{j,2}_{OCV,k}+V^{j,3}_{OCV,k}\\
    R^j_0 = R^{j,1}_0 +R^{j,2}_0+R^{j,3}_0 
\end{gather}
where $V_k=$ bus voltage, $N_p=$ number of cells in parallel, and $j=$ branch.

\subsubsection{Battery Degradation Model}
Degradation effects on Lipo battery parameters experiencing different aging mechanisms in lithium-ion cathode material are described in \cite{WOHLFAHRT2004}.  Battery aging results in power fade, capacity fade, and other degradation effects. Temperature effects on parameters in the ECS cell model are presented in \cite{plett2015batteryvol2} and are manifested as increased internal resistance. Service limits for sUAS batteries are defined in \cite{astm2014} as a pack that has lost $20\%$ of its rated capacity. Battery degradation models aim to determine which parameters and range of values to vary during simulation to estimate degraded battery performance. Degraded battery parameters are provided in Table \ref{tb:BattHealthParam}. This list of parameters is not exhaustive but provides a useful baseline.  Our simulations assume degradation/fault information will be available for our MDP since existing Battery Management Systems can flag faults \cite{weicker2013}. 

 \begin{table}[h]
    \caption{Battery degraded health parameter values.}
    \label{tb:BattHealthParam}
    \begin{center}
    \begin{tabular}{|c||c|}
    \hline 
    \textbf{Type} & \textbf{Description} \\
    \hline
    \textit{Capacity Fade}  & $0.8\times Q$  \\
    \hline
    \textit{Power Fade}  & $2\times R_0$ \\
    \hline
    \textit{Temperature effect}  & $R_{0L}:1.5\times R_0,$if $T<50F$ \\
    \hline
   \end{tabular}
    \end{center}
  \end{table}

\subsection{Motor Current Draw Model}
\begin{figure}[!htbp]
  \centering
    \includegraphics[width=0.3\textwidth]{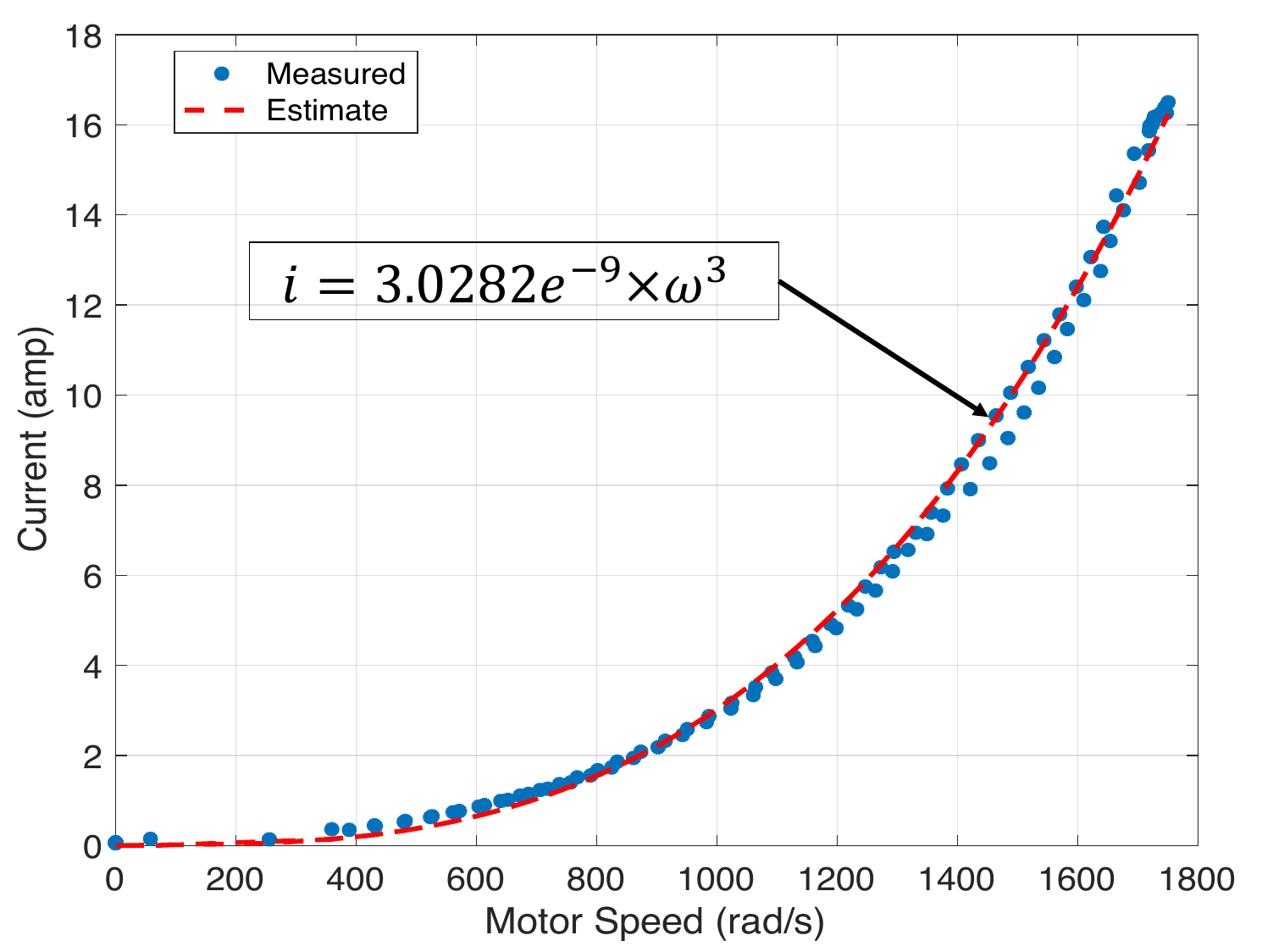}
    \caption{Motor current variation as a function of motor speed (RPM). }
    \label{fig:MotorCurrent}
\end{figure}

This section described the model used to determine the current draw from sUAS brushless DC motors (BLDC). Given our hexacopter case studies, rotors are numbered from 1-6 and are located at a distance $L$ from the sUAS center of gravity. A hexacopter's total motor thrust magnitude $T$ and torque magnitude $\tau_i$ about each body axis are related to six motor speeds as described by the following equations \cite{Prashin2019}: 

\begin{scriptsize}
\begin{align*}
\begin{pmatrix} \omega_1^2 \\ \omega_2^2 \\ \omega_3^2 \\ \omega_4^2 \\ \omega_5^2 \\ \omega_6^2 \end{pmatrix}
 = 
\setlength\arraycolsep{1.5pt}
\begin{pmatrix} C_T &C_T & C_T &C_T&C_T &C_T  \\ \frac{LC_T}{2} &  \frac{-LC_T}{2}& -LC_T & \frac{-LC_T}{2} & \frac{LC_T}{2} &LC_T\\ \frac{-\sqrt{3}lC_T}{2} & \frac{-\sqrt{3}lC_T}{2} & 0 &\frac{\sqrt{3}LC_T}{2} &\frac{\sqrt{3}lC_T}{2} & 0 \\ C_Q & -C_Q & C_Q & -C_Q &C_Q&-C_Q \end{pmatrix} ^\dagger
\begin{pmatrix}  T \\  \tau_x \\ \tau_y \\ \tau_z \end{pmatrix}
\label{eq:force}
\end{align*}
\end{scriptsize}
where $\omega_i=$ motor angular speed, $L=$ arm length (m), $C_T=$ thrust coefficient, $C_Q=$ torque coefficient, $\dagger=$ pseudoinverse.  A dynamometer\footnote{https://www.rcbenchmark.com/pages/series-1580-thrust-stand-dynamometer} was used to collect data for full throttle cycle of a BLDC reference motor. Least-squares curve fitting was performed to determine the current drawn by a motor for a given angular speed in our hexacopter experiments. Figure \ref{fig:MotorCurrent} shows the curve fit equation.

\subsection{Battery MDP Model Formulation}
Battery reconfiguration is modelled as a stationary infinite-horizon Markov Decision Process \cite{Martin2005} with the assumption that the system is fully observable. An MDP is defined by tuple $\langle T,S,A_s,p(.|s,a),r(s,a) \rangle $, where $T$ is decision epoch, $S$ is the finite set of system states, $A_s$ is allowable actions for states $s\in S$, $p(.|s,a)$ is state transition probability tensor and $r(s,a)$ is the reward for executing action $a\in A_s$ in state $s\in S$. The MDP computes actions that maximize expected value for each state based on the Bellman equation. Classic algorithms such as Value Iteration or Policy Iteration can be used to determine optimal MDP policy $\pi^\star$. 

 \begin{table}[h]
    \caption{Batttery MDP state definition.}
    \label{tb:MDPState}
    \begin{center}
    \begin{tabular}{|c||c|}
    \hline 
    \textbf{States,i-(1,2)} & \textbf{Criteria} \\
    \hline
    \textit{$B_iS_1$}  & EOD $>$ RFD +$t_{sf}$  \\
    \hline
    \textit{$B_iS_2$}  & RFD $>$ EOD $>$ RFD+$t_{sf}$ \\
    \hline
    \textit{$B_iS_3$}  & EOD $<$ RFD \\
    \hline
    \textit{$B_iC_0$} & Cell Voltage $\geq$ Critical Voltage Value(3.4VDC) \\
    \hline 
    \textit{$B_iC_1$}   &Cell Voltage $<$ Critical Voltage Value \\ 
    \hline
    \textit{$B_iSw$} & Battery Switch Position ie. ON or OFF  \\ 
    \hline
    \textit{$I$},Motor Current & $I_H:I>0.2I_{max}$, $I_L:I<0.2I_{max}$ \\
    \hline
     $FAILURE$   & Cell voltage $\leq$ 3.3VDC and is unusable \\ 
    \hline
    \end{tabular}
    \end{center}
    where, EOD = End of Discharge (time), RFD = Remaining Flight Duration (time), $t_{sf}=$ Safety Margin (time), $I_{max}=105$ amps
  \end{table}

\subsubsection{MDP State Definition}
Battery MDP states describing cell voltage, mission success status, and battery switch positions are defined per Table \ref{tb:MDPState} as state vector $[I,B_1Sw, B_1S_{1,2,3}, B_1C_{0,1},B_2Sw, B_2S_{1,2,3}, B_2C_{0,1}]$ and a singleton $FAILURE$ state.  $B_iS_{1,2,3}$ is dependent on an inequality relation of EOD and RFD. When the battery is in $B_iS_{1,2}$ there is sufficient energy to complete the mission. However, when the battery reaches $B_iS_3$ the battery will deplete before the mission ends. The element $B_iC_{0,1}$ indicates cell voltage and informs the decision-maker to take appropriate action so that the battery voltage does not reach a disabling limit. $B_iSw$ provides information on which battery of series-parallel battery pack is in use. The elements of battery MDP state are computed from the minimum (worst-case) value of any cell in each series battery pack.  The $I_{max}$ value for state $I$  is determined from the maximum permissible battery current draw per manufacturer data. The special state $FAILURE$ occurs when cell voltage dips below the lowest Lipo voltage limit, i.e. $3.3VDC$. In low voltage conditions, electronic speed controllers (ESC) used to drive BLDC motors will shut down resulting in a loss of thrust and subsequent sUAS uncontrolled descent.

Battery health state is captured in the discrete MDP state values listed in Table \ref{tb:Condition}. Because we do not model battery health transition dynamics in this work we assume conditions are known or accurately measured and remain constant over each flight simulation.  

  \begin{table}[h]
    \caption{Battery health state values.}
    \label{tb:Condition}
    \begin{center}
    \begin{tabular}{|c||c|}
    \hline 
    \textbf{Condition,i-(1,2)} & \textbf{Criteria} \\
    \hline
    \textit{$B_iF_1$}  & Healthy battery  \\
    \hline
    \textit{$B_iF_2$}  & Medium health battery  \\
    \hline
    \textit{$B_iF_3$}  & Unhealthy battery \\
    \hline
    \textit{$T$},Temperature & $T_H:T>50F$, $T_L:T<50F$ \\
    \hline
    \end{tabular}
    \end{center}
    where a Medium health battery has either power or capacity fade in at least one cell and an Unhealthy Battery has power and capacity fade in at least one cells.  T = ambient temperature.
  \end{table}
\subsubsection{Battery MDP Action Definitions}
Battery MDP actions are defined in Table \ref{tb:Actions} for the series-parallel battery pack shown in Figure \ref{fig:series-parallel}. 
  
  \begin{table}[h]
    \caption{Battery management MDP action definitions.}
    \label{tb:Actions}
    \begin{center}
    \begin{tabular}{|c||c|}
    \hline 
    \textbf{Actions} & \textbf{Description} \\
    \hline
    \textit{$UseBatt_1$}  & $Sw_1$ is ON , $Sw_2$ is OFF  \\
    \hline
    \textit{$UseBatt_2$}  & $Sw_1$ is OFF , $Sw_2$ is ON   \\
    \hline
    \textit{$UseBoth$}  & $Sw_1$ is ON , $Sw_2$ is ON  \\
    \hline
   \end{tabular}
    \end{center}
  \end{table}
  
\subsubsection{Reward Definition}
The reward assigned to each MDP state/action pair is defined by the following equations:
\begin{gather}
     R(s,UseBatt_1) = w_1R_{B_1S_*}+w_2R_{B_1C_*}-w_3R_{B_2Sw} \\
        \label{eq:reward1}\\
    \begin{split}
    R(s,UseBoth)=w_1\frac{\sum^2_{i=1}R_{B_iS_*}}{2}+ w_2\frac{\sum^2_{i=1}R_{B_iC_*}}{2} -  \\
    w_3 \left[\frac{\sum^2_{i=1}R_{B_iSw}}{2}-1 \right]   
    \end{split}\\
    R_{B_iS_*} =
    \begin{cases}
      0 & \text{For all conditions with $B_iS_1$}\\
    [-5,-10] & \text{Based on relative conditions of}\\   &\text{the two battery packs in $B_iS_2$}\\
     [-20,-25]& \text{Based on relative conditions of}\\   &\text{the two battery packs in $B_iS_3$}\\
      \end{cases} \\
     R_{B_iC_*} =
    \begin{cases}
      0 & \text{For $C_*=C_0$}\\
    -10 & \text{For $C_*=C_1$}\\   
      \end{cases} \\
      R_{B_iSw} =
    \begin{cases}
      1 & \text{For $B_iSw=ON$}\\
      0 & \text{For $B_iSw=OFF$}\\   
      \end{cases} \\
     R(Failure)= -30 \\
     w_1+w_2+w_3 = 1 \\
\end{gather}

where $i=[1,2]$. Reward $R(s,UseBatt_2)$ is computed analogously to Eqn. \ref{eq:reward1} depicting $R(s,UseBatt_1)$. 
Reward component $R_{B_iS_\star}$ penalizes states with lower EOD value and poor health relative to the other battery.  $R_{B_1C_\star}$ penalizes states with low voltage value, while $R_{B_1Sw}$ penalizes a switch position action that cycles the batteries (unnecessarily). The $FAILURE$ state it assigned the highest penalty. The different reward components are weighted with user-specified $w_i$.

\subsubsection{Policy Implementation}
\begin{figure}[t!]
  \centering
    \includegraphics[width=0.5\textwidth]{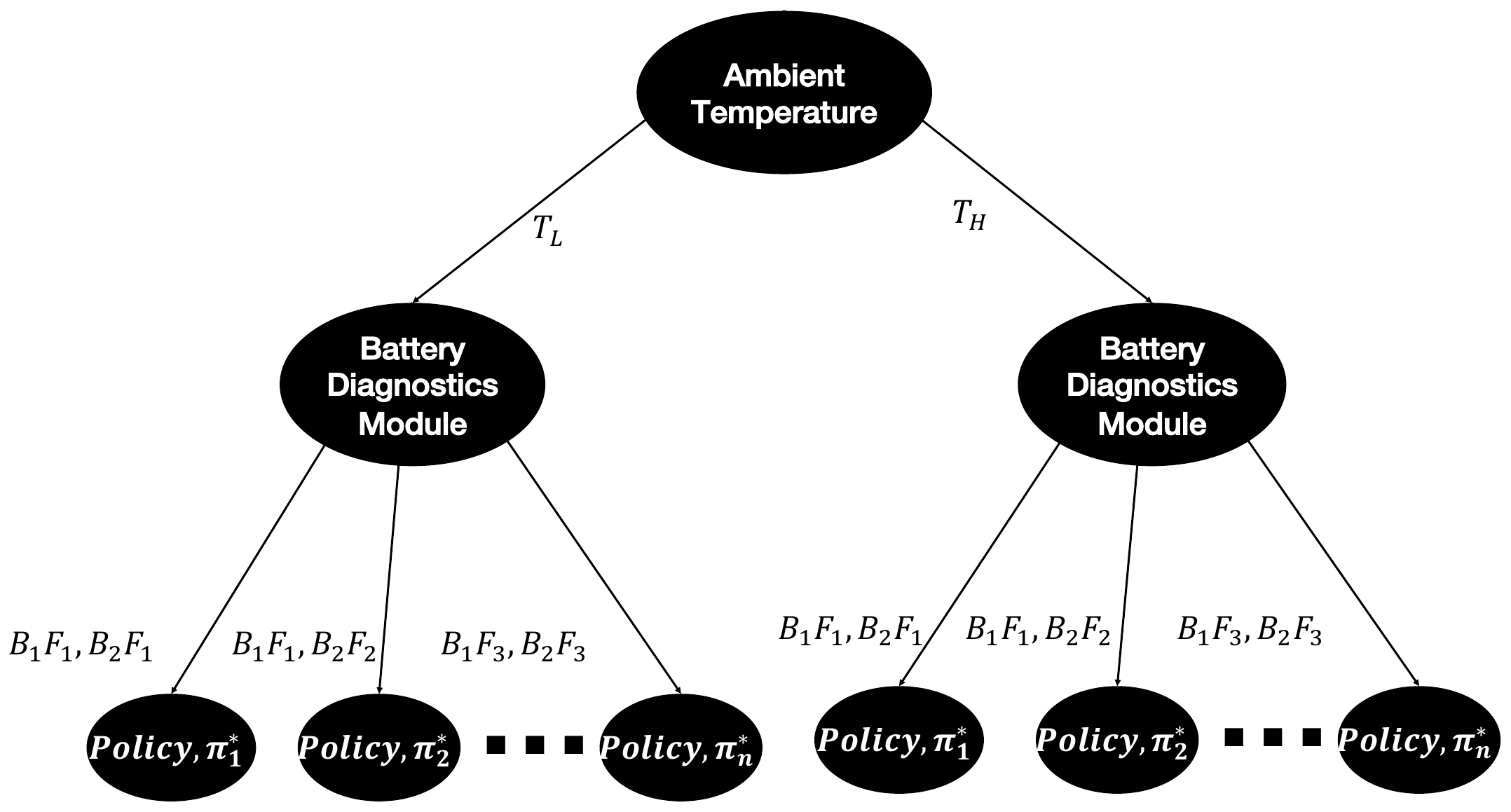}
    \caption{MDP decision tree to match policy $\pi^*$ with current battery health conditions. }
    \label{fig:Tree}
\end{figure}

Each optimal policy $\pi^\star$ for the MDP was determined using Value Iteration with a discount factor $\gamma=0.95$. Policies were stored in a lookup table for different conditions and selected as shown in Figure \ref{fig:Tree}.

\section{Monte Carlo Simulation}
\begin{figure*}[t!]
        \centering
        \includegraphics[width = 5.85in , height=2.55in ]{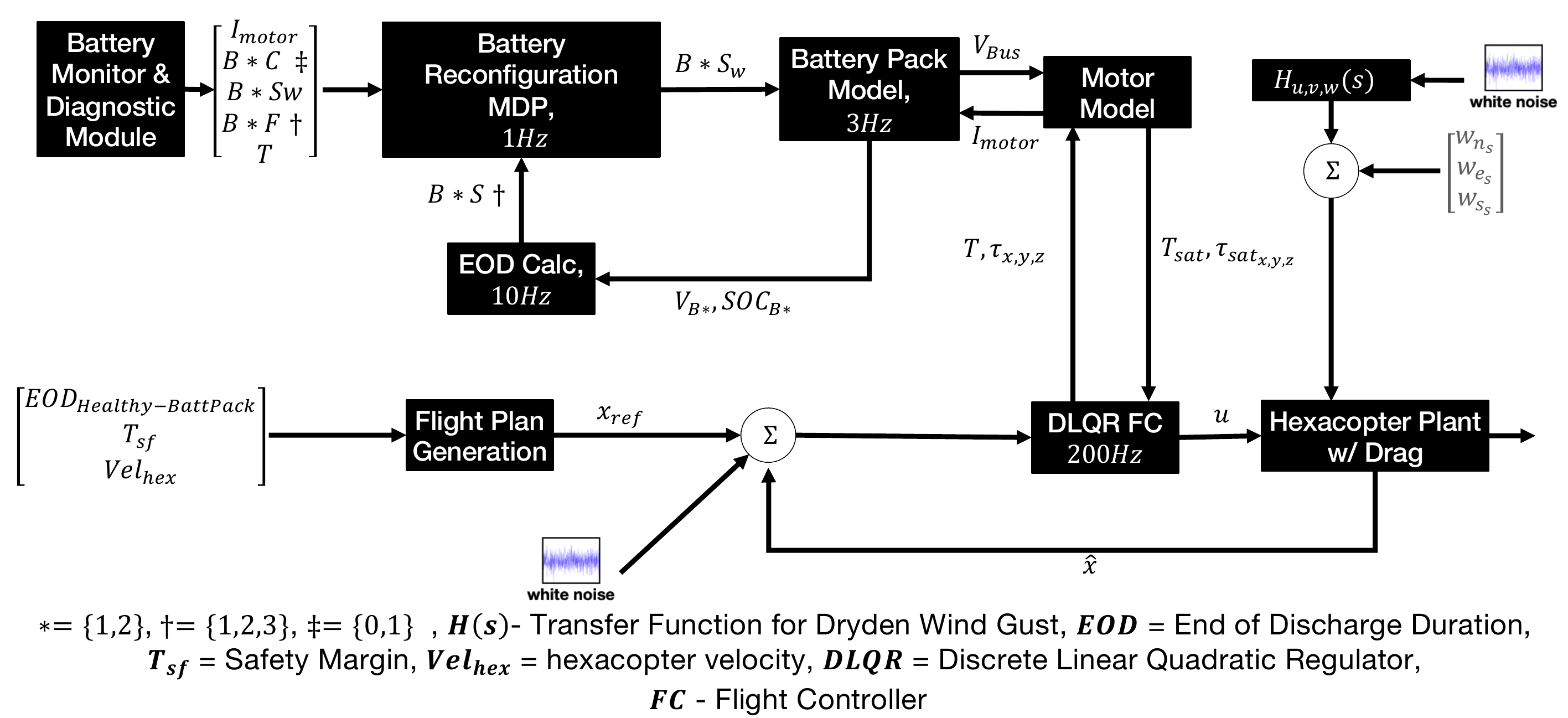}
        \caption{Monte Carlo simulation system diagram.}
        \label{fig:SimArch}
    \end{figure*}

\label{Sec:MCSim}
Monte Carlo (MC) simulations are used to determine the battery MDP state transition probabilities. The list of parameters varied for MC simulations is presented in Table \ref{tb:MCSimParam}. The range for initial cell voltages and motor speeds are based on past flight tests \cite{Prashin2019} conducted with a hexacopter. 
  
 \begin{table}[h]
    \caption{Parameters varied for Monte Carlo simulations.}
    \label{tb:MCSimParam}
    \begin{center}
    \begin{tabular}{|c||c|}
    \hline 
    \textbf{Parameter} & \textbf{Range} \\
    \hline
    \textit{Action}  & $UseBatt_1,UseBatt_2,UseBoth$  \\
    \hline
    \textit{Safety Margin,$t_{sf}$}  & [5,10]  \\
    \hline
    \textit{Initial Cell Voltage,$B_{i,VDC}$}  & [4.05,4.12]$VDC$ \\
    \hline
    \textit{Hexacopter Velocity,$Vel_{hex}$}  & [1,9]$m/s$ \\
    \hline
    \textit{Wind Speed,$w$}  & [1,3]$m/s$ \\
    \hline
    \textit{Wind Direction,$\theta$}  & [0,$\pi$]$rad$ \\
    \hline
    \textit{$Wind Gust^\ddagger$ }  & [Low Light, Low Moderate, \\                & Medium Light, Medium Moderate]    \\
    \hline
    \textit{Cell Health,$B_{i}F_*$}  & [$F_1,F_2,F_3$] \\
    \hline
   \end{tabular}
    \end{center}
    $\ddagger=$ Dryden wind model\cite{Beard2012}
  \end{table}
Data flow through Monte Carlo system simulations is shown in Figure \ref{fig:SimArch}. The first step is randomly assigning MC parameters from the prescribed ranges and generating a representative flight plan for a package drop mission. For trajectory tracking, a Discrete Linear Quadratic Regulator (DLQR) state feedback controller is implemented with 200Hz update rate. Closed-loop control forces and torques are fed into the motor model to determine current, scaling by battery bus voltage, and any motor thrust saturation condition. Motor current draw is considered the dominant source for battery voltage drop. Hotel load i.e., current draw from an onboard computer, communication system, and sensors, is considered relatively small. Current drawn is used to determine battery EOD time. The same system diagram is used to determine state transition probabilities and simulate MDP policy execution in this work. The battery MDP policy executor responsible for battery management actions runs at 1Hz.  
\begin{figure}[t!]
        \includegraphics[width = 3.5in , height=2.65in ]{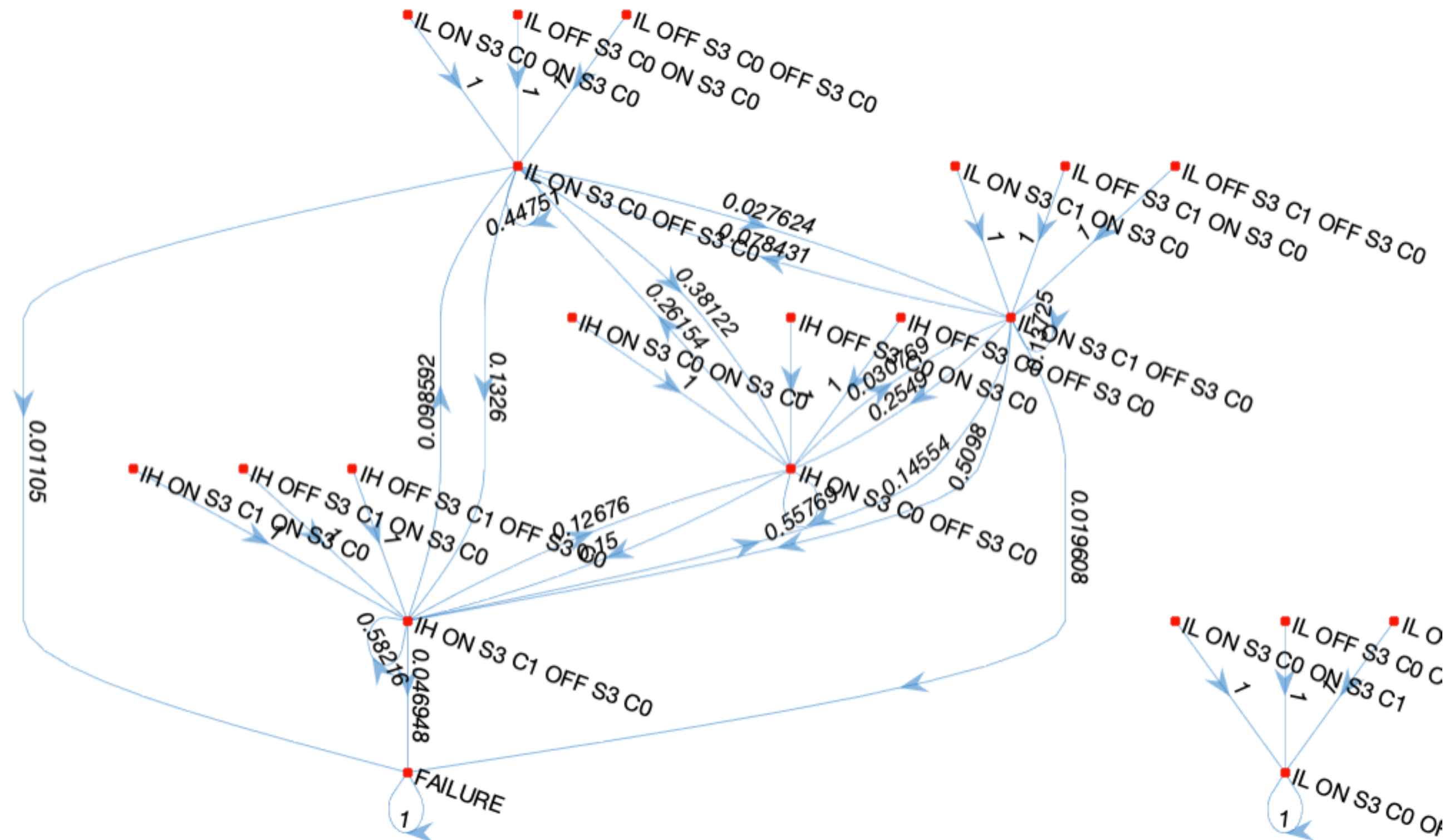}
        \caption{Directed graph showing battery MDP state transitions for action $UseBatt_1$. Nodes labeled  IL\textcolor{red}{OFFS3C0}\textcolor{blue}{OFFS3C0} describe states associated with \textcolor{red}{$Batt_1$} and \textcolor{blue}{$Batt_2$}.  }
        \label{fig:StateTransition}
    \end{figure}

An example of state transition probabilities obtained from 5000 MC simulations is shown in Figure \ref{fig:StateTransition}. We assume each commanded battery switch action works with 100\% reliability in this work.  Further, we assume a battery switch occurs instantaneously such that there is no intermediate change in any other MDP state features, i.e. temperature, current, health, or battery state.


\section{Results}
\label{Sec:Results}
A series of case studies were performed to determine the performance of series-parallel battery pack usage with the MDP policy benchmarked against a fixed battery usage rule. Case studies are summarized in Table \ref{tb:Case}.
 \begin{table}[h]
    \caption{sUAS battery case study summary.}
    \label{tb:Case}
    \begin{center}
    \begin{tabular}{|c||c||c||c|}
    \hline 
    \textbf{No.} & Flight Time(sec) & \textbf{Action} & \textbf{Battery Health} \\
    \hline
    1 & $EOD_{Batt_1}$ &$UseBoth$  & $B_1F_{1,2,3}$, $B_2F_{1,2,3}$\\
    \hline
    2& $EOD_{Batt_1}$ & $\pi^\star$  & $B_1F_{1,2,3}$, $B_2F_{1,2,3}$\\
    \hline
    3& $EOD_{BothBatt}$ & $UseBoth$  & $B_1F_{1,2,3}$, $B_2F_{1,2,3}$\\
    \hline
    4& $EOD_{BothBatt}$ & $\pi^\star$  & $B_1F_{1,2,3}$, $B_2F_{1,2,3}$\\
    \hline
   \end{tabular}
    \end{center}
  \end{table}
  
\begin{figure}[t!]
        \includegraphics[width = 2.85in , height=2.25in ]{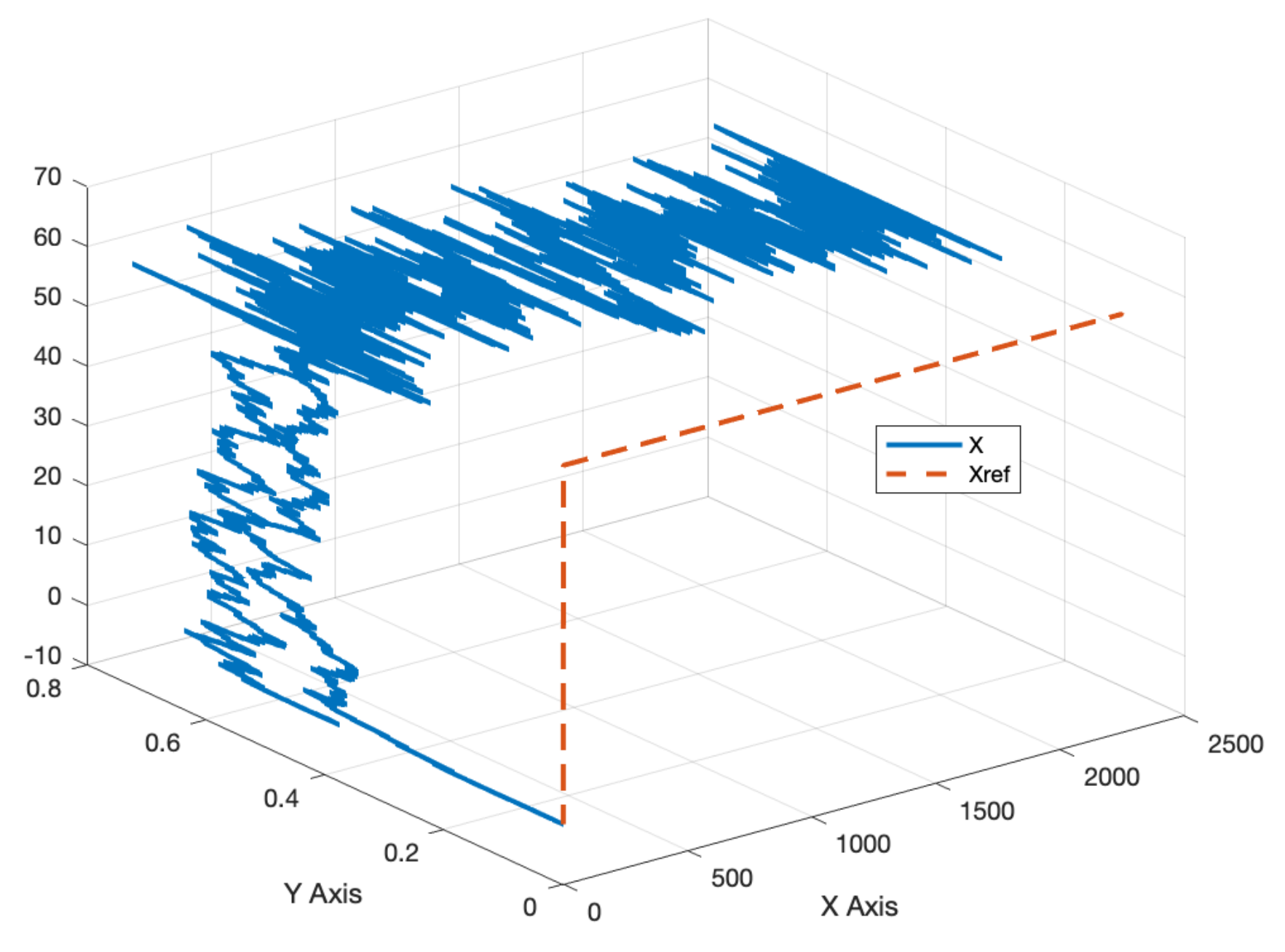}
        \caption{3-D flight trajectory executed by the simulated hexacopter.}
        \label{fig:3Dtraj}
\end{figure}

\begin{figure}[t!]
    \includegraphics[width = 3.0in , height=3.0in ]{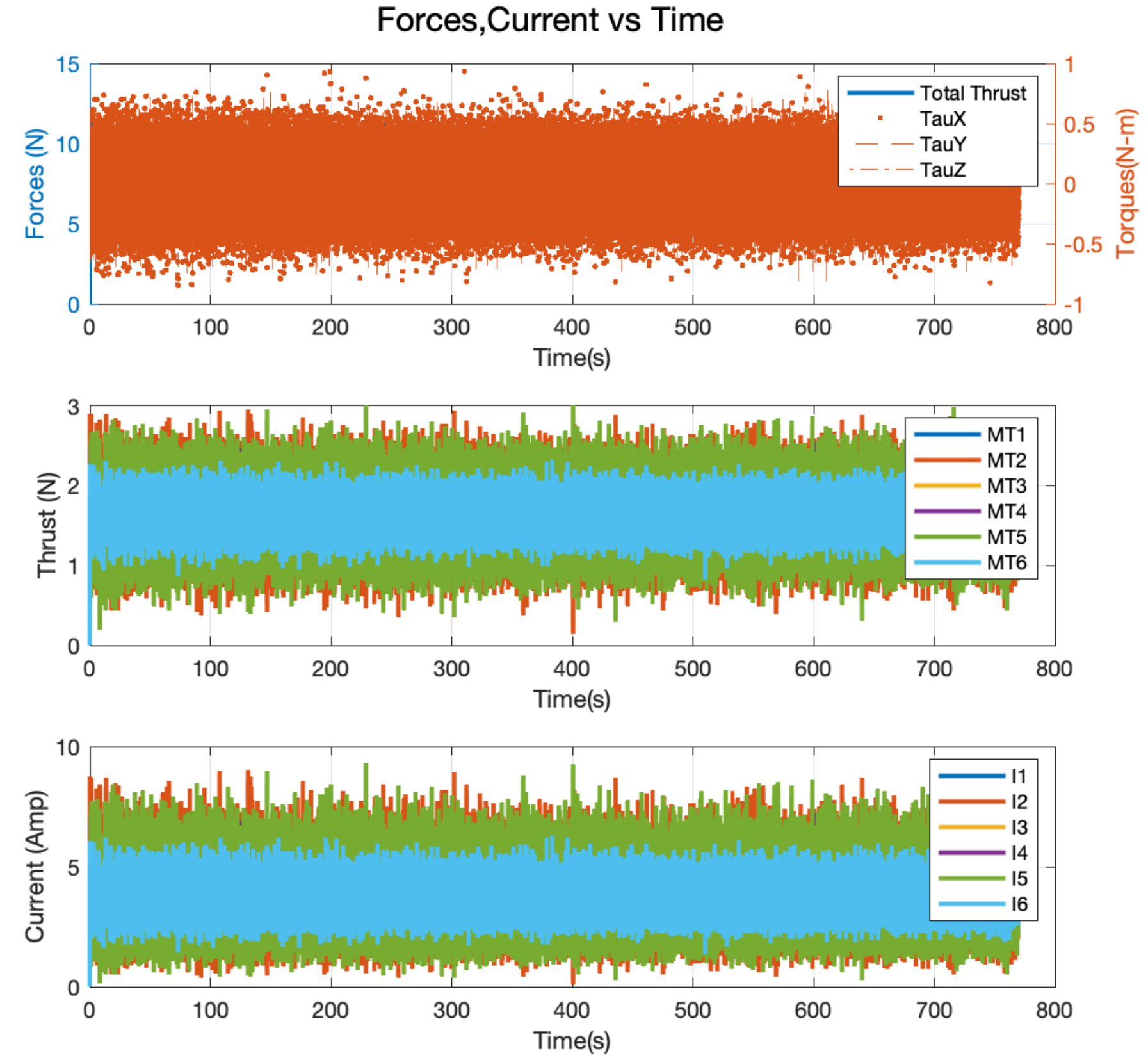}
    \caption{Vehicle forces, motor thrust and current draw during a hexacopter trajectory tracking simulation.}
    \label{fig:Forces}
\end{figure}

In studies 1 and 2, battery pack EOD time is twice the flight time, so all mission simulations were successful. 
3D trajectory tracking of the simulated hexacopter is shown in Figure \ref{fig:3Dtraj}. The hexacopter climbs to a known altitude, translates to its destination, hovers for a specified time to simulate package drop, then returns to the start location. Since no integrator term was added to the controller, there is an offset between reference and actual trajectory. The controller could have been tuned to provide better performance; however, noise is an important and practical element of the simulations. 
Force, torque and motor current time responses are plotted in Figure \ref{fig:Forces}.  White noise is introduced into the system so data appears noisy to emulate non-ideal real world conditions. 

\begin{figure*}[htb!]
    \centering
    \begin{subfigure}[t]{0.5\textwidth}
        \centering
        \includegraphics[height=2.2in]{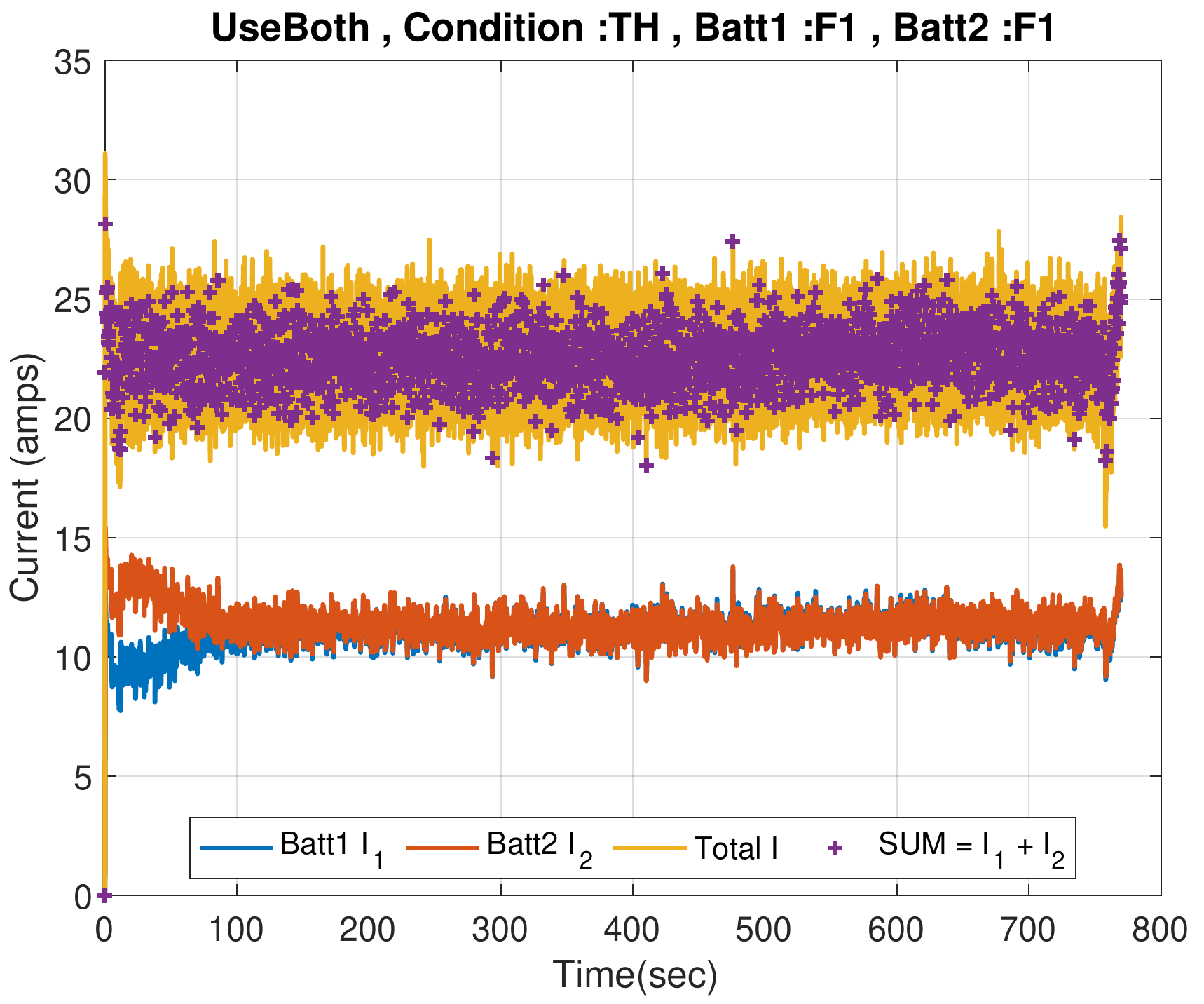}
        \caption{Total current and current passing through each branch of a series-parallel battery pack. }
    	\label{fig:UseBothITHF1F1}
    \end{subfigure}%
    \begin{subfigure}[t]{0.5\textwidth}
        \centering
        \includegraphics[height=2.2in]{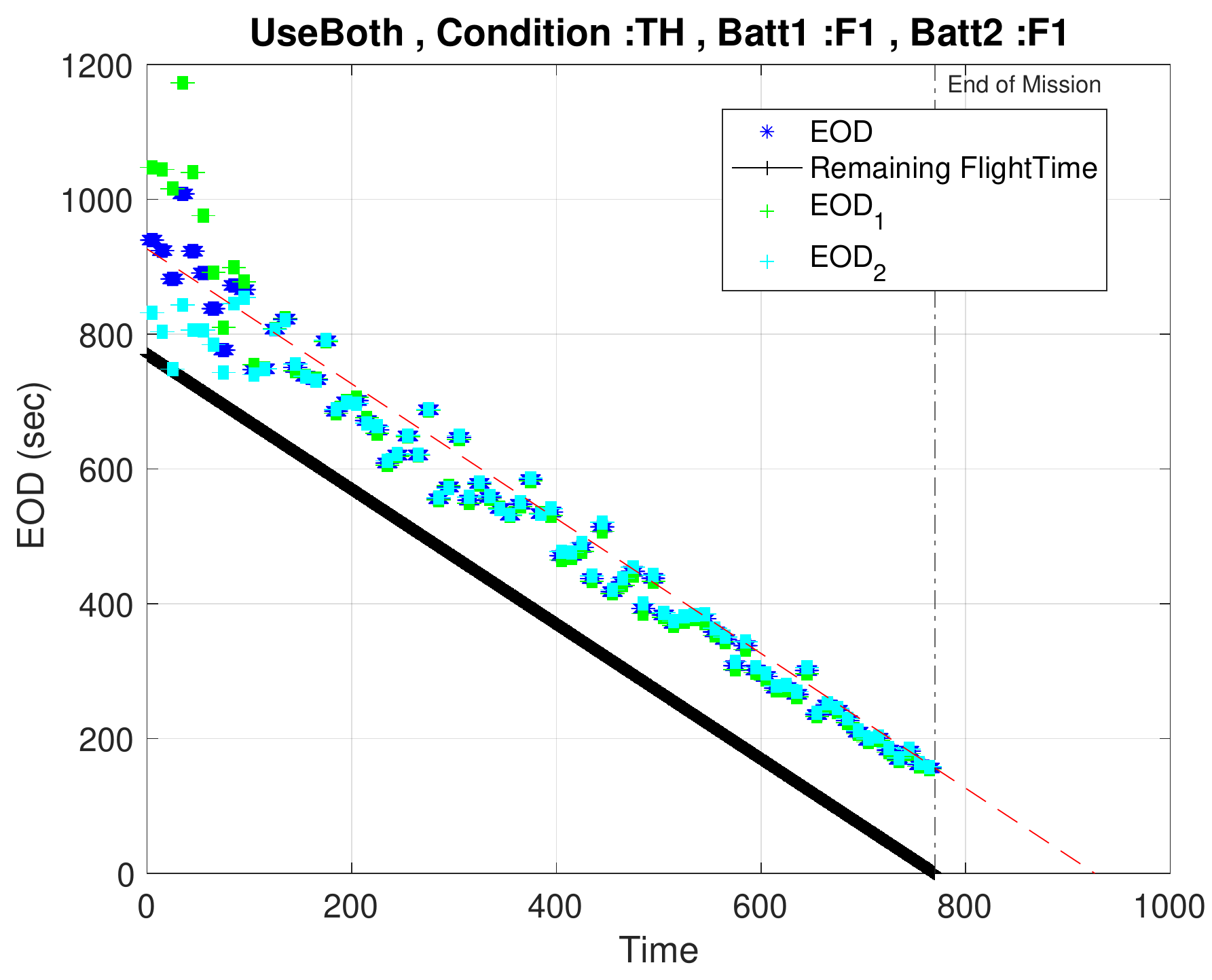}
        \caption{EOD variation.}
    	\label{fig:UseBothEODTHF1F1}
    \end{subfigure}
    \caption{Simulated battery data for Case Study 3 Scenario 1 where all  cells of $Batt_1$ and $Batt_2$ are healthy ($F_1$).}
    \label{fig:THF1F1}
\end{figure*}

In case study 3, three scenarios were considered based on battery health: ${(B_1F_1, B_2F_2),(B_1F_1, B_2F_3),(B_1F_3, B_2F_3)}$. Results from scenario one i.e. $(B_1F_1, B_2F_2)$ are presented in Figure \ref{fig:THF1F1}. Battery current consumed is plotted in Figure \ref{fig:UseBothITHF1F1}. A slight difference in current drawn from the two battery branches exists until $100 sec$ into the flight due to an initial difference in internal resistance. Once there is an unequal voltage drop, the current draw equalizes. The EOD for both the batteries is shown in Figure \ref{fig:UseBothEODTHF1F1}.The thick black line defines remaining flight time, and the vertical dashed-dotted line represents end of mission. Since both the batteries are healthy, the mission is completed successfully. 

\begin{figure*}[htb!]
    \centering
    \begin{subfigure}[t]{0.5\textwidth}
        \centering
        \includegraphics[height=2.2in]{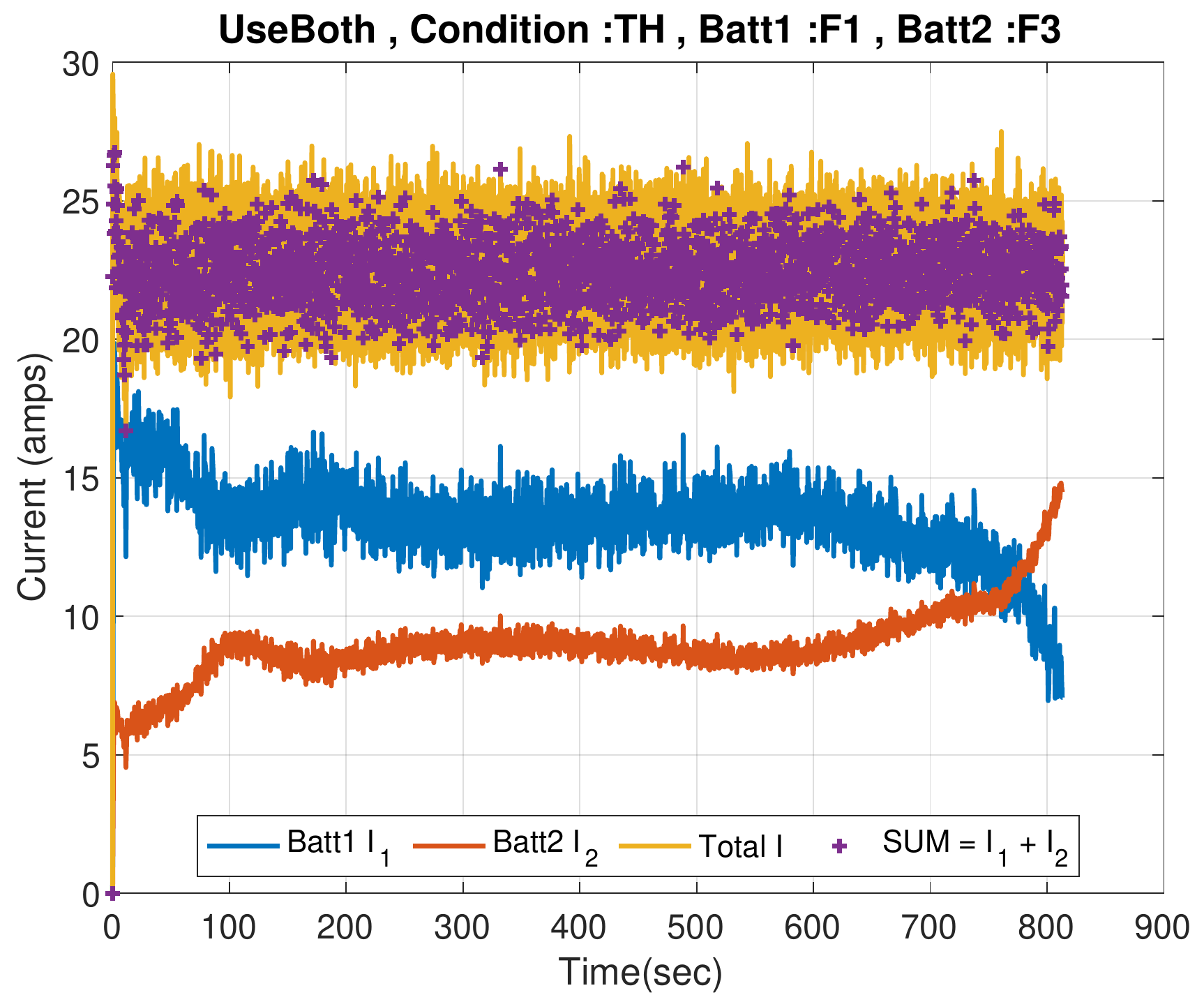}     
        \caption{Total current and current passing through each branch of a series-parallel battery pack. }
    	\label{fig:UseBothITHF1F3}
    \end{subfigure}%
    \begin{subfigure}[t]{0.5\textwidth}
        \centering
        \includegraphics[height=2.2in]{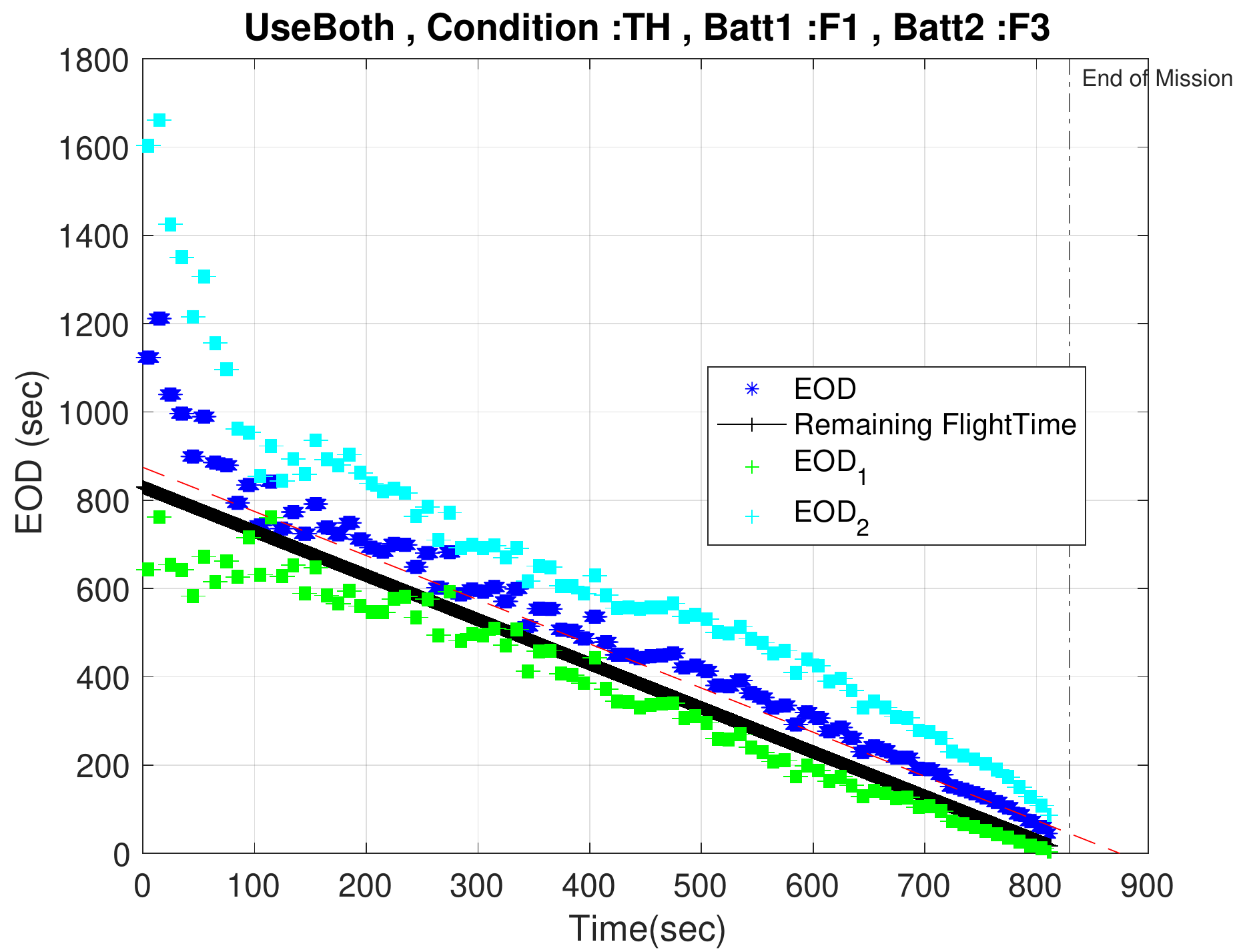}
        \caption{EOD variation.}
    	\label{fig:UseBothEODTHF1F3}
    \end{subfigure}
    \caption{Simulated battery data for Case 3 Scenario 2 where all cells of $Batt_1$ are healthy but cells of $Batt_2$ encounter both capacity and power fade (i.e. $F_3$). }
    \label{fig:THF1F3}
\end{figure*}

Results from scenario two i.e.$(B_1F_1, B_2F_3)$ of case study 3 are plotted in Figure \ref{fig:THF1F3}. Due to difference in health of the parallel batteries there is a significant difference in current drawn from the battery branches as seen in Figure \ref{fig:UseBothITHF1F3}. Current draw in parallel cells is modeled per \cite{Gong2014}. Individual battery pack variations are shown in Figure \ref{fig:UseBothEODTHF1F3}. Inspection of EODs at $\sim 810 sec$ shows that the sUAS fails to complete the mission due to imbalance in branch currents. Even though $Batt_1$ is healthy, its $EOD_1$ value is always below the Remaining Flight Time line. In such scenarios mission success cannot be guaranteed.

\begin{figure*}[htb!]
    \centering
    \begin{subfigure}[t]{0.5\textwidth}
        \centering
        \includegraphics[height=2.2in]{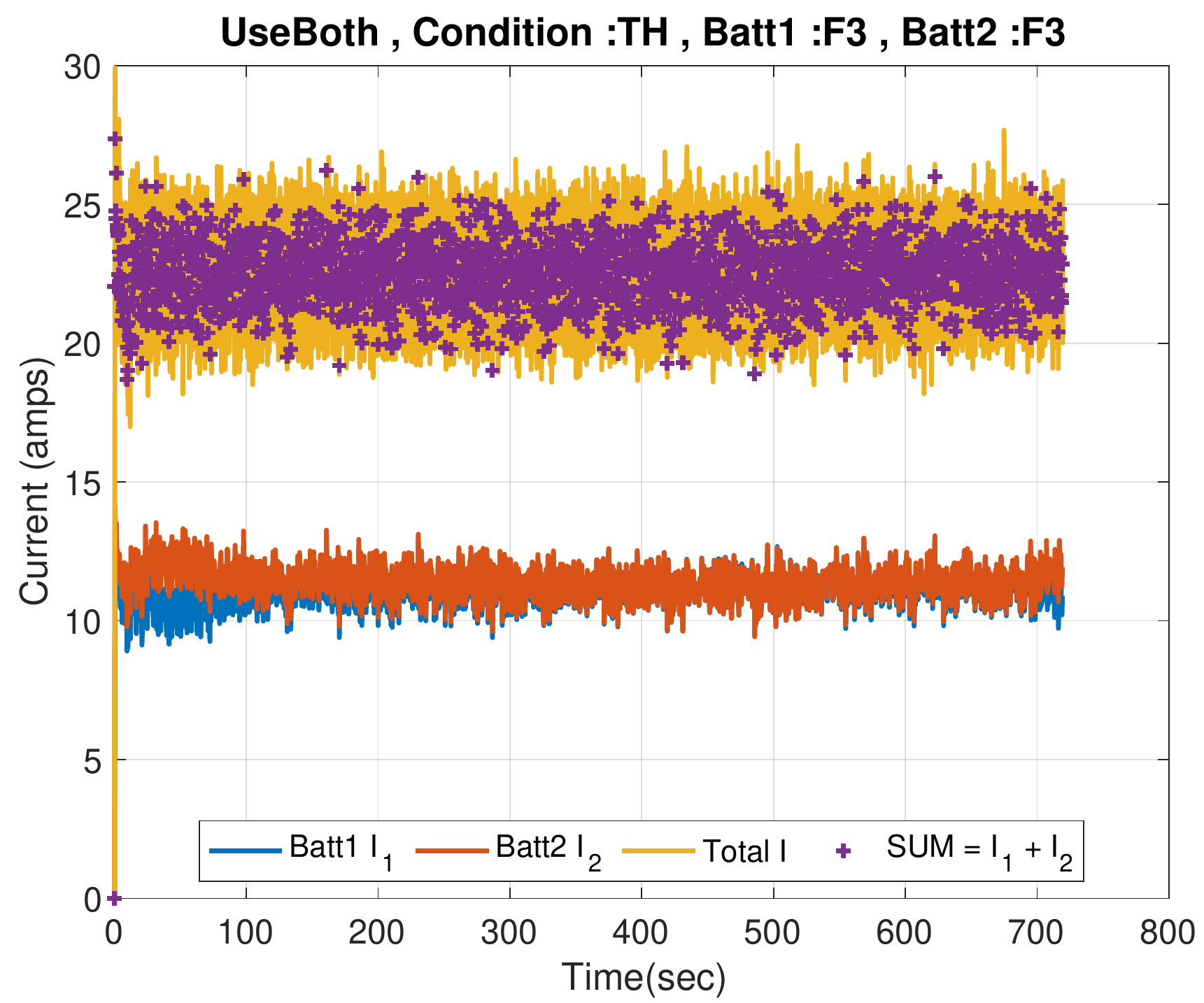}     
        \caption{Total current and current passing through each branch of the series-parallel battery pack. }
    	\label{fig:UseBothITHF3F3}
    \end{subfigure}%
    \begin{subfigure}[t]{0.5\textwidth}
        \centering
        \includegraphics[height=2.2in]{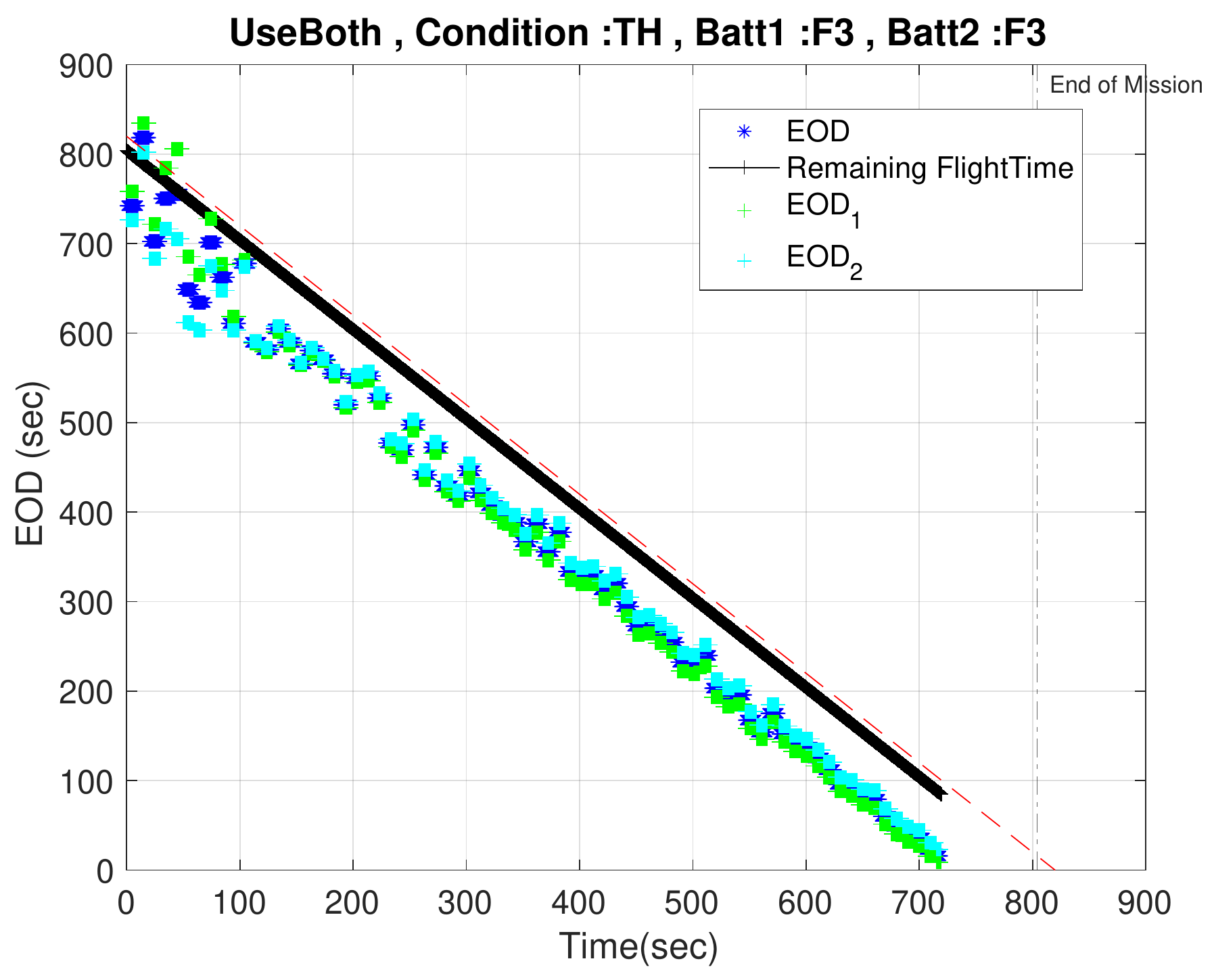}
        \caption{EOD variation.}
    	\label{fig:UseBothEODTHF3F3}
    \end{subfigure}
    \caption{Simulated battery data for Case Study 3 Scenario 3 where all cells of $Batt_1$ and all cells of $Batt_2$ experience both capacity and power fade (i.e. $F_3$). }
    \label{fig:THF3F3}
\end{figure*}

In scenario three i.e. i.e.$(B_1F_3, B_2F_3)$ of case study 3, both of the batteries are in poor health due to aging.  The current drawn through each branch is thus about the same as shown in Figure \ref{fig:UseBothITHF3F3}. However, since both batteries have capacity fade, the EOD value for both batteries is below the remaining flight time line throughout the mission, which implies the mission will fail. Such a situation highlights the importance of a higher-level planner that redefines the mission or executes an appropriate contingency plan when insufficient battery energy remains to safely continue the original flight plan. 

\begin{figure*}[htbp]
    \centering
    \begin{subfigure}[t]{0.5\textwidth}
        \centering
        \includegraphics[height=2.2in]{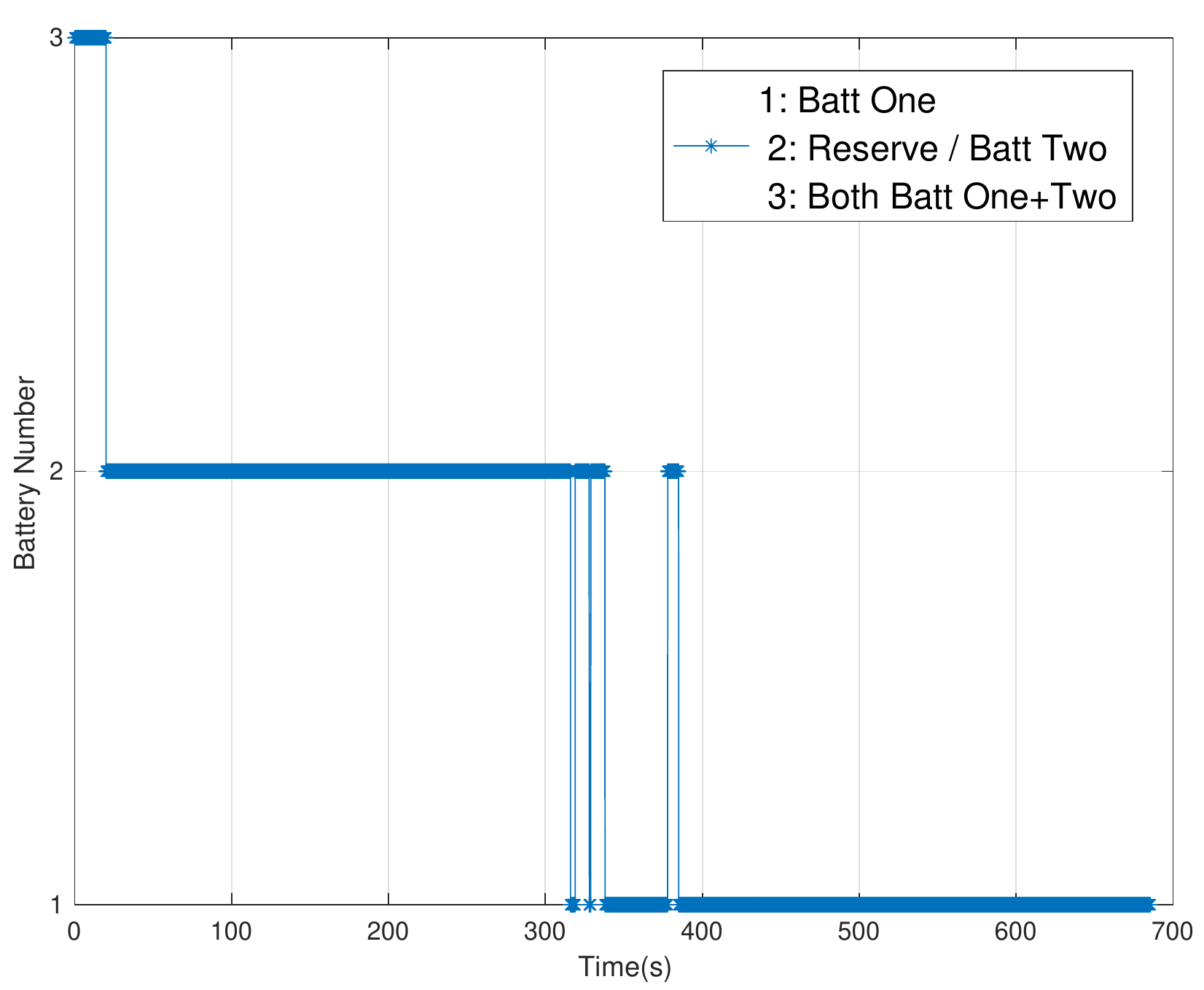}     
        \caption{MDP Policy battery switching actions for the series-parallel battery configuration. }
    	\label{fig:PolicySwitch}
    \end{subfigure}%
    \begin{subfigure}[t]{0.5\textwidth}
        \centering
        \includegraphics[height=2.2in]{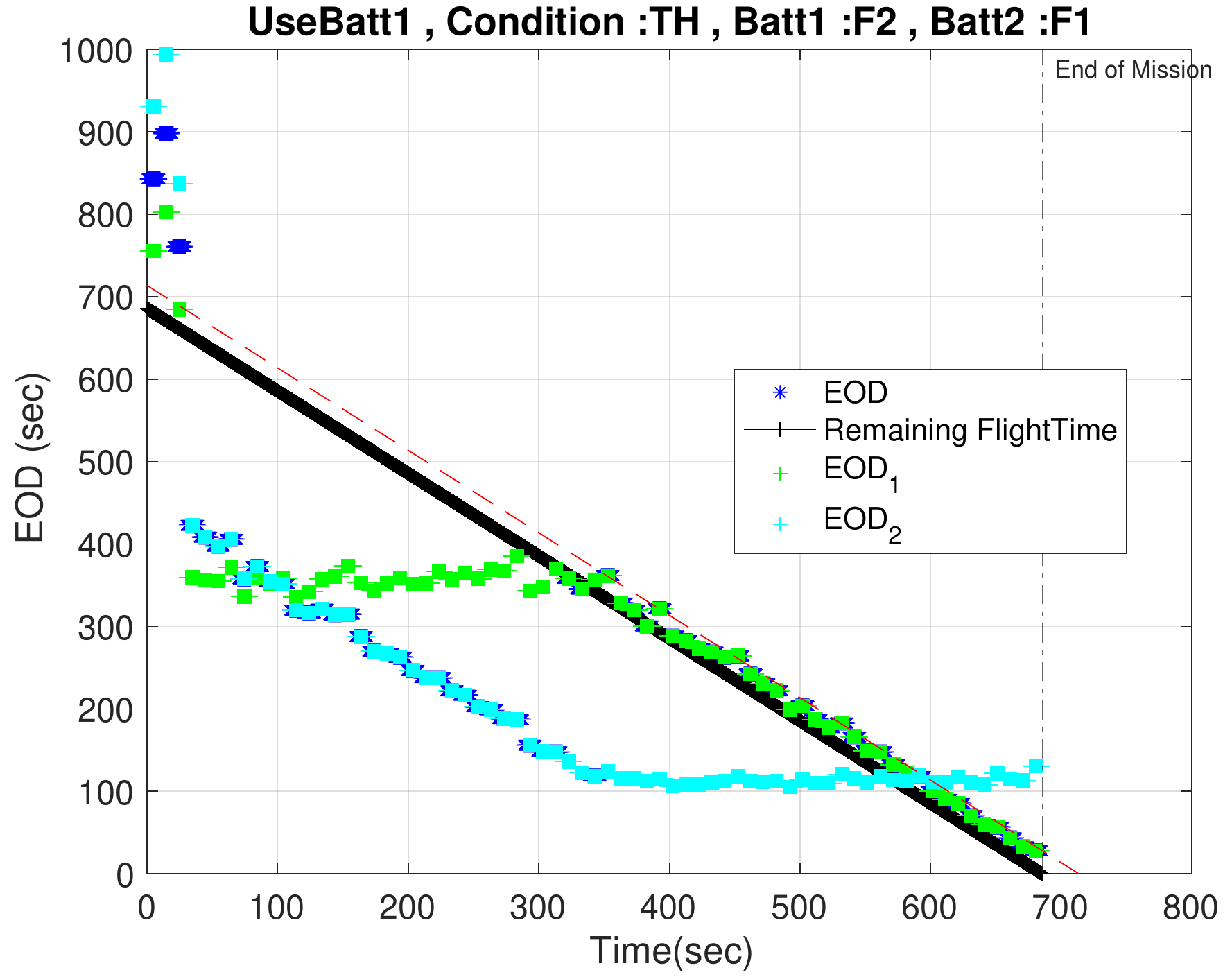}
        \caption{EOD Variation.}
    	\label{fig:PolicyEOD}
    \end{subfigure}
    \caption{Analysis of simulated trajectory tracking data for Case Study 4, Scenario 1 where $Batt_1$ has one cell with capacity fade and $Batt_2$ is healthy. }
    \label{fig:Policy}
\end{figure*}

For case study 4, results from scenario $(B_1F_2, B_2F_1)$ are plotted in Figure \ref{fig:Policy}. The MDP policy switches with $UseBoth\rightarrow UseBatt_2$ occurs at $\sim 20sec$ and not to $UseBatt_1$ due to its poor health. Preference in this case is given to utilizing a single battery instead of both batteries when both are in the $B_iS_1$ state. With this formulation, repeated battery switching can occur.  The UAS was still able to complete the mission as shown in Figure \ref{fig:PolicyEOD}. A similar scenario can be set up with a single action of $UseBoth$. An observed benefit of battery reconfiguration is that an unused battery can rest for subsequent use as needed. 

\section{Conclusion}
\label{Sec:Conclusion}
This paper has presented an approach to sUAS battery management with an MDP. With our state-space abstraction, MDP reward tuning required substantial effort to achieve desirable behaviour. Our proposed battery reconfiguration MDP provides benefits in term of optimal battery switching. However, additional statistical analysis is required to determine the efficacy of the implemented policy. As seen from baseline case studies, if both batteries have unexpectedly poor health flight the original flight plan can still be impossible to achieve. Battery prognostics results must therefore be shared with a flight planner capable of updating the flight plan in real-time to assure the sUAS lands before battery energy is fully expended. 

\bibliographystyle{IEEEtran}
\bibliography{main.bbl}

\begin{thebibliography}{10}
\providecommand{\url}[1]{#1}
\csname url@samestyle\endcsname
\providecommand{\newblock}{\relax}
\providecommand{\bibinfo}[2]{#2}
\providecommand{\BIBentrySTDinterwordspacing}{\spaceskip=0pt\relax}
\providecommand{\BIBentryALTinterwordstretchfactor}{4}
\providecommand{\BIBentryALTinterwordspacing}{\spaceskip=\fontdimen2\font plus
\BIBentryALTinterwordstretchfactor\fontdimen3\font minus
  \fontdimen4\font\relax}
\providecommand{\BIBforeignlanguage}[2]{{%
\expandafter\ifx\csname l@#1\endcsname\relax
\typeout{** WARNING: IEEEtran.bst: No hyphenation pattern has been}%
\typeout{** loaded for the language `#1'. Using the pattern for}%
\typeout{** the default language instead.}%
\else
\language=\csname l@#1\endcsname
\fi
#2}}
\providecommand{\BIBdecl}{\relax}
\BIBdecl

\bibitem{Wing}
\BIBentryALTinterwordspacing
``Learn about how wing delivery works,'' \emph{Wing (Accessed 2021)}. [Online].
  Available: \url{https://wing.com/about-delivery/}
\BIBentrySTDinterwordspacing

\bibitem{Flytrex}
\BIBentryALTinterwordspacing
``Faq,'' \emph{Flytrex (Accessed 2021)}. [Online]. Available:
  \url{https://www.flytrex.com/support/faq/}
\BIBentrySTDinterwordspacing

\bibitem{Zipline}
\BIBentryALTinterwordspacing
``How it works,'' \emph{Zipline (Accessed 2021)}. [Online]. Available:
  \url{https://flyzipline.com/how-it-works/}
\BIBentrySTDinterwordspacing

\bibitem{Osborne2019}
\BIBentryALTinterwordspacing
M.~{Osborne}, J.~{Lantair}, Z.~{Shafiq}, X.~{Zhao}, V.~{Robu}, D.~{Flynn}, and
  J.~{Perry}, ``{UAS Operators Safety and Reliability Survey: Emerging
  Technologies towards the Certification of Autonomous UAS},'' \emph{4th
  International Conference on System Reliability and Safety (ICSRS)}, pp.
  203--212, 2019. [Online]. Available:
  \url{https://doi.org/10.1109/ICSRS48664.2019.8987692}
\BIBentrySTDinterwordspacing

\bibitem{chetan2019health}
\BIBentryALTinterwordspacing
C.~{Kulkarni} and M.{Corbetta}, ``Health management and prognostics for
  electric aircraft powertrain,'' \emph{AIAA Propulsion and Energy Forum},
  2019. [Online]. Available:
  \url{https://arc.aiaa.org/doi/abs/10.2514/6.2019-4474}
\BIBentrySTDinterwordspacing

\bibitem{schacht2018prognosis}
\BIBentryALTinterwordspacing
R.{Schacht-Rodr{\'\i}guez}, J-C.{Ponsart}, CD.{Garc{\'\i}a-Beltr{\'a}n}, and
  CM.{Astorga-Zaragoza}, ``Prognosis and health management for the prediction
  of uav flight endurance,'' \emph{10th IFAC Symposium on Fault Detection,
  Supervision and Safety for Technical Processes SAFEPROCESS}, vol.~51, no.~24,
  pp. 983--990, 2018. [Online]. Available:
  \url{https://doi.org/10.1016/j.ifacol.2018.09.705}
\BIBentrySTDinterwordspacing

\bibitem{hogge2018verification}
E.F.{Hogge}, B.~{Bole}, S.~{Vazquez}, C.~{Kulkarni}, T.H.{Strom}, B.L.{Hill},
  K.M.{Smalling}, and C.C.{Quach}, ``Verification of prognostic algorithms to
  predict remaining flying time for electric unmanned vehicles,''
  \emph{International Journal of Prognostics and Health Management}, 2018.

\bibitem{ELEFTHEROGLOU2019}
\BIBentryALTinterwordspacing
N.{Eleftheroglou}, S.~{Mansouri}, T.{Loutas}, P.{Karvelis}, G.{Georgoulas},
  G.{Nikolakopoulos}, and D.{ Zarouchas}, ``Intelligent data-driven prognostic
  methodologies for the real-time remaining useful life until the
  end-of-discharge estimation of the lithium-polymer batteries of unmanned
  aerial vehicles with uncertainty quantification,'' \emph{Applied Energy},
  vol. 254, 2019. [Online]. Available:
  \url{https://doi.org/10.1016/j.apenergy.2019.113677}
\BIBentrySTDinterwordspacing

\bibitem{balaban2013modeling}
E.{Balaban} and J.J.{Alonso}, ``A modeling framework for prognostic decision
  making and its application to uav mission planning,'' in \emph{Annual
  Conference of The Prognostics and Health Management Society}, 2013, pp.
  1--12.

\bibitem{balaban2013development}
E.~{Balaban}, S.{Narasimhan}, M.{Daigle}, I.{Roychoudhury}, A.{Sweet},
  C.{Bond}, and G.{Gorospe}, ``Development of a mobile robot test platform and
  methods for validation of prognostics-enabled decision making algorithms,''
  \emph{International Journal of Prognostics and Health Management}, vol.~4,
  no.~1, p.~87, 2013.

\bibitem{schacht2019}
\BIBentryALTinterwordspacing
R.~{Schacht-Rodríguez}, J.~C. {Ponsart}, C.~D. {García-Beltrán}, C.~M.
  {Astorga-Zaragoza}, and D.~{Theilliol}, ``Mission planning strategy for
  multirotor uav based on flight endurance estimation*,'' in
  \emph{International Conference on Unmanned Aircraft Systems (ICUAS)}, June
  2019, pp. 778--786. [Online]. Available:
  \url{https://doi.org/10.1109/ICUAS.2019.8798292}
\BIBentrySTDinterwordspacing

\bibitem{tang2008}
\BIBentryALTinterwordspacing
L.~{Tang}, G.~J. {Kacprzynski}, K.~{Goebel}, A.~{Saxena}, B.~{Saha}, and
  G.~{Vachtsevanos}, ``Prognostics-enhanced automated contingency management
  for advanced autonomous systems,'' in \emph{International Conference on
  Prognostics and Health Management}, 2008, pp. 1--9. [Online]. Available:
  \url{https:/doi/org/10.1109/PHM.2008.4711448}
\BIBentrySTDinterwordspacing

\bibitem{tang2010}
\BIBentryALTinterwordspacing
L.~{Tang}, G.~J. {Kacprzynski}, K.~{Goebel}, and G.~{Vachtsevanos}, ``Case
  studies for prognostics-enhanced automated contingency management for
  aircraft systems,'' in \emph{IEEE Aerospace Conference}, March 2010, pp.
  1--11. [Online]. Available: \url{https://doi.org/10.1109/AERO.2010.5446844}
\BIBentrySTDinterwordspacing

\bibitem{zhang2014}
B.~{Zhang}, L.{Tang}, J.{Decastro}, M.{Roemer}, and K.~{Goebel}, ``Autonomous
  vehicle battery state-of-charge prognostics enhanced mission planning,''
  \emph{International Journal of Prognostics and Health Management}, vol.~5,
  no.~8, 2014.

\bibitem{Poteiger2017}
\BIBentryALTinterwordspacing
T.~{Potteiger}, W.~{Strayhorn}, K.~R. {Pence}, and G.~{Karsai}, ``A dependable,
  prognostics-incorporated, n-s modular battery reconfiguration scheme with an
  application to electric aircraft,'' \emph{IEEE/AIAA 36th Digital Avionics
  Systems Conference}, pp. 1--9, 2017. [Online]. Available:
  \url{https://doi.org/10.1109/DASC.2017.8102033}
\BIBentrySTDinterwordspacing

\bibitem{Ci2016}
\BIBentryALTinterwordspacing
S.~{Ci}, N.~{Lin}, and D.~{Wu}, ``Reconfigurable battery techniques and
  systems: A survey,'' \emph{IEEE Access}, vol.~4, pp. 1175--1189, 2016.
  [Online]. Available: \url{https://doi.org/10.1109/ACCESS.2016.2545338}
\BIBentrySTDinterwordspacing

\bibitem{Jain2020}
\BIBentryALTinterwordspacing
K.~P. {Jain} and M.~W. {Mueller}, ``Flying batteries: In-flight battery
  switching to increase multirotor flight time,'' in \emph{IEEE International
  Conference on Robotics and Automation (ICRA)}, 2020, pp. 3510--3516.
  [Online]. Available: \url{https://doi.org/10.1109/ICRA40945.2020.9197580}
\BIBentrySTDinterwordspacing

\bibitem{IVERSEN20141}
\BIBentryALTinterwordspacing
E.~{Iversen}, J.M.{Morales}, and H.{Madsen}, ``Optimal charging of an electric
  vehicle using a markov decision process,'' \emph{Applied Energy}, vol. 123,
  pp. 1--12, 2014. [Online]. Available:
  \url{https://doi.org/10.1016/j.apenergy.2014.02.003}
\BIBentrySTDinterwordspacing

\bibitem{THEIN2014142}
\BIBentryALTinterwordspacing
S.{Thein} and Y.S.{Chang}, ``Decision making model for lifecycle assessment of
  lithium-ion battery for electric vehicle - a case study for smart electric
  bus project in korea,'' \emph{Journal of Power Sources}, vol. 249, pp.
  142--147, 2014. [Online]. Available:
  \url{https://doi.org/10.1016/j.jpowsour.2013.10.078}
\BIBentrySTDinterwordspacing

\bibitem{Mikolajczak2011}
\BIBentryALTinterwordspacing
C.~{Mikolajczak}, M.{Kahn}, K.{White}, and R.T.{Long}, ``Lithium-ion batteries
  hazard and use assessment,'' \emph{Fire Protection Research Foundation
  Report}, 2011. [Online]. Available:
  \url{https://www.nfpa.org/News-and-Research/Data-research-and-tools/Hazardous-Materials/Lithium-ion-batteries-hazard-and-use-assessment}
\BIBentrySTDinterwordspacing

\bibitem{Michael2018}
\BIBentryALTinterwordspacing
M.~{Logan}, J.{Gundlach}, and T.~{Vranas}, ``Design considerations for safer
  small uas,'' \emph{AIAA Information Systems- AIAA Infotech @ Aerospace},
  2018. [Online]. Available:
  \url{https://arc.aiaa.org/doi/abs/10.2514/6.2018-1724}
\BIBentrySTDinterwordspacing

\bibitem{Prashin2020}
P.{Sharma} and E.M.{Atkins}, ``Prognostics-based decision making for safe
  autonomous flight,'' \emph{Doctoral Symposium, Annual Conference of the PHM
  Society 2020}, 2020.

\bibitem{plett2015battery}
G.~{Plett}, \emph{Battery Management Systems, Volume I: Battery
  modeling}.\hskip 1em plus 0.5em minus 0.4em\relax Artech House, 2015, vol.~1.

\bibitem{saxena2018}
\BIBentryALTinterwordspacing
S.~{Saxena}, Y.~{Xing}, and M.~{Pecht}, ``{PHM} of li-ion batteries,''
  \emph{Prognostics and Health Management of Electronics}, pp. 349--375, 2018.
  [Online]. Available: \url{https://doi.org/10.1002/9781119515326.ch13}
\BIBentrySTDinterwordspacing

\bibitem{WOHLFAHRT2004}
M.~{Wohlfahrt-Mehrens}, C.~{Vogler}, and J.~{Garche}, ``Aging mechanisms of
  lithium cathode materials,'' \emph{Journal of Power Sources}, vol. 127,
  no.~1, pp. 58--64, 2004.

\bibitem{plett2015batteryvol2}
G.~{Plett}, \emph{Battery Management Systems, Volume II: Equivalent Circuit
  Methods}.\hskip 1em plus 0.5em minus 0.4em\relax Artech House, 2015, vol.~1,
  ch.~1.

\bibitem{astm2014}
\BIBentryALTinterwordspacing
ASTM-F3005-14a, ``Standard specification for batteries for use in small
  unmanned aircraft systems (suas),'' \emph{ASTM International,West
  Conshohocken, PA, 2014}, 2014. [Online]. Available:
  \url{https://doi.org/10.1520/F3005-14A}
\BIBentrySTDinterwordspacing

\bibitem{weicker2013}
P.~{Weicker}, \emph{A Systems Approach to Lithium-Ion Battery
  Management}.\hskip 1em plus 0.5em minus 0.4em\relax Artech House, 2013,
  ch.~1.

\bibitem{Prashin2019}
\BIBentryALTinterwordspacing
P.{Sharma} and E.M.{Atkins}, ``Experimental investigation of tractor and pusher
  hexacopter performance,'' \emph{Journal of Aircraft}, vol.~56, no.~5, pp.
  1920--1934, 2019. [Online]. Available:
  \url{https://doi.org/10.2514/1.C035319}
\BIBentrySTDinterwordspacing

\bibitem{Martin2005}
\BIBentryALTinterwordspacing
M.L.{Putterman}, \emph{Markov Decision Processes: Discrete Stochastic Dynamic
  Programming}.\hskip 1em plus 0.5em minus 0.4em\relax John Wiley \& Sons,
  2005. [Online]. Available: \url{https://doi.org/10.1002/9780470316887}
\BIBentrySTDinterwordspacing

\bibitem{Beard2012}
R.~{Beard} and T.{McLain}, \emph{Small Unmanned Aircraft: Theory and
  Practice}.\hskip 1em plus 0.5em minus 0.4em\relax Princeton University Press,
  2012, ch.~4.

\bibitem{Gong2014}
\BIBentryALTinterwordspacing
X.~{Gong}, R.~{Xiong}, and C.~C. {Mi}, ``Study of the characteristics of
  battery packs in electric vehicles with parallel-connected lithium-ion
  battery cells,'' \emph{IEEE Applied Power Electronics Conference and
  Exposition}, pp. 3218--3224, 2014. [Online]. Available:
  \url{https://doi.org/10.1109/APEC.2014.6803766}
\BIBentrySTDinterwordspacing

\end{thebibliography}

\end{document}